\documentclass[10pt,twocolumn,letterpaper]{article}

\usepackage{wacv}
\usepackage{times}
\usepackage{epsfig}
\usepackage{graphicx}
\usepackage{amsmath}
\usepackage{amssymb}

\usepackage{algpseudocode}
\usepackage{booktabs}
\usepackage{capt-of}
\usepackage{color}
\usepackage{etoolbox}
\usepackage{fixmath}
\usepackage[numbers,sort]{natbib}
\usepackage{siunitx}

\def\wacvPaperID{874} %

\wacvfinalcopy %

\ifwacvfinal
\def\assignedStartPage{1} %
\fi

\ifwacvfinal
\usepackage[breaklinks=true,bookmarks=false]{hyperref}
\else
\usepackage[pagebackref=true,breaklinks=true,colorlinks,bookmarks=false]{hyperref}
\fi

\MakeRobust{\Call}
\robustify\bfseries
\allowdisplaybreaks

\ifwacvfinal
\else
\fi

\newcommand{\PAR}[1]{\smallskip \noindent \textbf{#1}}
\newcommand{\PARbegin}[1]{\noindent \textbf{#1}}

\newcommand{\figref}[1]{Figure~\ref{#1}}
\newcommand{\tabref}[1]{Table~\ref{#1}}
\renewcommand{\eqref}[1]{Eq.~(\ref{#1})}

\renewcommand{\vec}[1]{\mathbold{#1}}
\newcommand{\mat}[1]{\mathbold{#1}}

\begin{document}

\title{Learning to Reconstruct 3D Non-Cuboid Room Layout\\from a Single RGB Image}

\author{
Cheng Yang\textsuperscript{\rm 1,2}\thanks{Equal contributions.} \thanks{This work was done when Yang Cheng was a student at UESTC.} \quad
Jia Zheng\textsuperscript{\rm 3}$^*$ \quad
Xili Dai\textsuperscript{\rm 1,4} \quad
Rui Tang\textsuperscript{\rm 3} \quad
Yi Ma\textsuperscript{\rm 4} \quad
Xiaojun Yuan\textsuperscript{\rm 1}\thanks{Corresponding author.} \\
\textsuperscript{\rm 1}University of Electronic Science and Technology of China \quad
\textsuperscript{\rm 2}Hikvision Research Institute \\
\textsuperscript{\rm 3}Manycore Tech Inc. (Kujiale) \quad
\textsuperscript{\rm 4}University of California, Berkeley \\
{\small \url{https://github.com/CYang0515/NonCuboidRoom}}
}

\maketitle

\begin{abstract}
Single-image room layout reconstruction aims to reconstruct the enclosed 3D structure of a room from a single image. Most previous work relies on the cuboid shape prior. This paper considers a more general indoor assumption, i.e., the room layout consists of a single ceiling, a single floor, and several vertical walls. To this end, we first employ Convolutional Neural Networks to detect planes and vertical lines between adjacent walls. Meanwhile, estimating the 3D parameters for each plane. Then, a simple yet effective geometric reasoning method is adopted to achieve room layout reconstruction. Furthermore, we optimize the 3D plane parameters to reconstruct a geometrically consistent room layout between planes and lines. The experimental results on public datasets validate the effectiveness and efficiency of our method.
\end{abstract}

\section{Introduction}

\begin{figure}[t]
    \centering
	\setlength{\tabcolsep}{1pt}
	\begin{tabular}{cc}
		\includegraphics[width=0.45\linewidth]{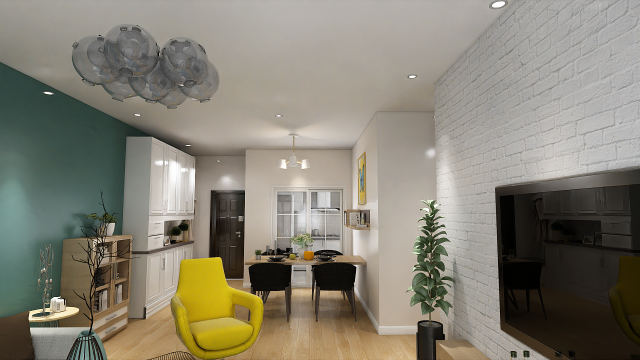} &
		\includegraphics[width=0.45\linewidth]{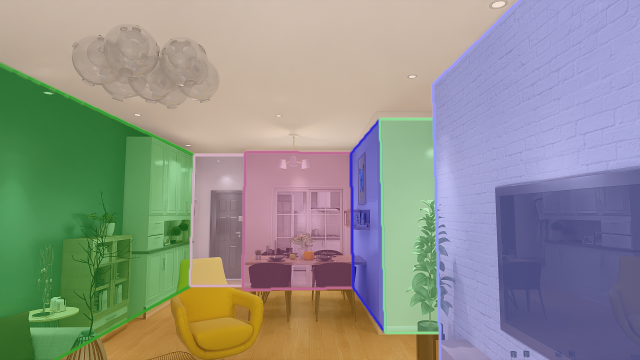}
		\tabularnewline
		{Input image} & {2D room layout}
		\tabularnewline
		\includegraphics[width=0.45\linewidth]{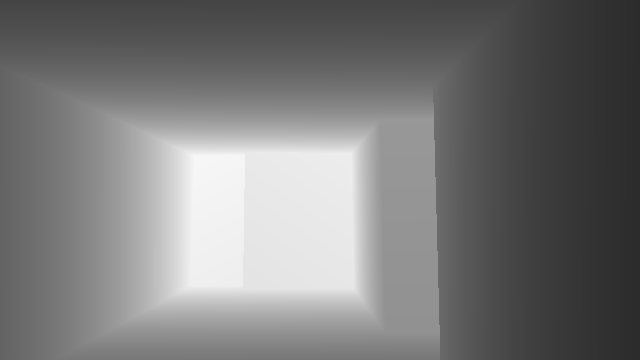} &
		\includegraphics[width=0.45\linewidth]{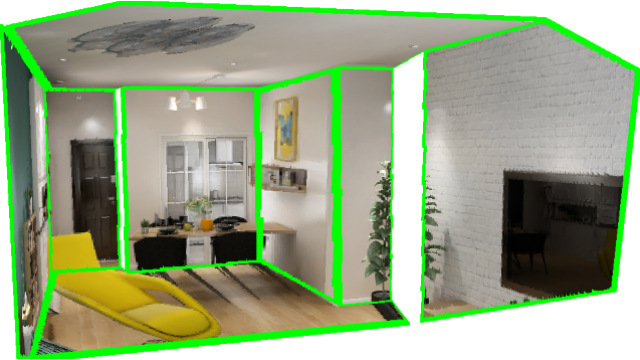}
		\tabularnewline
		{Depth map} & {3D room model}
		\tabularnewline
	\end{tabular}
	\caption{This paper tackles the room layout reconstruction from a single RGB image {\em without} cuboid-shape prior or Manhattan World assumption.}
	\label{fig:teaser}
\end{figure}

Room layout estimation aims to reconstruct the enclosed structure of an indoor scene, consisting of walls, ceiling, and floor (\figref{fig:teaser}). Estimating a 3D room layout from a single image plays a vital role in many applications such as robotics, Virtual Reality (VR), and Augmented Reality (AR).

In an early learning-based attempt, \citet{HedauHF09} assume a simple cuboid model for the room structure (ceiling, floor, and three walls). \citet{LSUN16} further define \num{11} types of cuboid-shape layouts to cover most of the possible situations under typical camera poses. Almost all existing methods~\cite{HedauHF09, SchwingHPU12, ZhangKSU13, SchwingFPU13, MallyaL15, DasguptaFCS16, RenLCK16, ZhaoLYGCZ17, LeeBMR17, HirzerRL20} follow these two cuboid-based definitions~\cite{HedauHF09, LSUN16} of the room layout, thereby being not flexible enough to handle variations in the real-world scenario.

Recent approaches~\cite{HowardJLP18, StekovicFL20} attempt to relax these assumptions by casting room layout estimation as a plane detection problem. For example, Planar R-CNN~\cite{HowardJLP18} modifies Faster R-CNN~\cite{RenHGS15} to detects 3D planes and Render-and-Compare (in short, RaC)~\cite{StekovicFL20} builds upon the advanced plane detection method PlaneRCNN~\cite{LiuKGFK19}. To correctly infer room layout with relaxed assumptions, two core challenges must be addressed properly. One challenge is how to infer the connectivity relations between planes in 3D space. Planar R-CNN~\cite{HowardJLP18} does not reason such relation and only reconstructs the piece-wise planar surfaces. RaC~\cite{StekovicFL20} defines a constrained discrete optimization problem to reason the relations. However, the optimization step is computationally expensive and maybe less robust due to the hand-crafted heuristics. The other challenge is how to deal with the occlusions, \ie, two adjacent walls in 2D space may be physically disconnected in 3D space. Planar R-CNN~\cite{HowardJLP18} directly uses a bounding box to locate the boundary of each plane coarsely. RaC~\cite{StekovicFL20} adopts RANSAC algorithm to fit the occlusion line to the points with the largest depth discrepancy changes in an analysis-by-synthesis fashion. This method requires several iterations, and the hyper-parameters of RANSAC algorithm should be carefully chosen.

Hence, to address the above challenges more effectively and efficiently, we assume that the room layout consists of a single ceiling, a single floor, and several vertical walls. Instead of only relying on 3D plane detection results, we introduce the 2D vertical lines of adjacent walls. This allows us to fully utilize the geometric relationship between planes and lines to solve the above challenges. Two adjacent walls in 2D image space are either physically connected or disconnected in 3D space: (i) Two physically connected walls form a vertical line segment in 2D image space. We can detect the line segment to adjust the 3D plane parameter estimations. (ii) Two physically disconnected walls form an occlusion line. We can directly detect the occlusion line to bound the planar surface.

To this end, we first train Convolutional Neural Networks (CNNs) to detect planes and vertical lines in the input RGB image. Meanwhile, we also estimate the 3D parameters (\ie, surface normal and offset) for each plane. Then, we explore the underlying geometric relationship between planes and lines to achieve room layout reconstruction. Specifically, we first calculate the intersection line of two adjacent walls and project it into 2D image space with the known camera intrinsic matrix. Depending on whether the projected intersection line lies between two adjacent walls or not, we classify the geometric relationship of these two walls as physically connected or disconnected in 3D space. Furthermore, if two adjacent walls are physically connected in 3D space, their projected intersection line should align with the corresponding detected line. Thus, we use the detected line as the geometric cues to optimize the 3D plane parameters, which enables our method to reconstruct a geometrically consistent 3D room layout. Otherwise, if two adjacent walls are not physically connected and the occlusion occurs, we directly use the corresponding detected line to separate them accurately.

In summary, \textbf{our contributions} are as follows: (i) We present a simple yet effective framework for 3D room layout reconstruction from a single RGB image. (ii) We propose to jointly detect 3D planes and vertical lines, and use vertical lines as complementary cues to assist the layout reconstruction. (iii) Experimental results on two challenging datasets, namely Structured3D dataset~\cite{ZhengZLTGZ20} and NYUv2 303 dataset~\cite{ZhangKSU13, SilbermanHKF12}, validate the effectiveness and efficiency of our method.

\section{Related Work}

\PARbegin{Room layout estimation.} In the literature, most existing methods tackle this problem with strict assumptions. \citet{LeeHK09} propose ``Indoor World'' model by combining the Manhattan World assumption and single-floor single-ceiling assumption. They could recover the 3D model by geometric reasoning on the configuration of edges. \citet{HedauHF09} propose to model the room by a parametric box (cuboid). They generate layout hypotheses by sampling rays from the detected vanishing points and then select the best layout hypothesis. The following methods~\cite{SchwingHPU12, ZhangKSU13, SchwingFPU13} follow this paradigm and improve this method.

Inspired by the recent success of CNN on semantic segmentation, several approaches train CNNs to classify pixels into boundaries~\cite{MallyaL15, RenLCK16, ZhaoLYGCZ17}, surfaces~\cite{DasguptaFCS16}, or corners~\cite{LeeBMR17}. Recently, several approaches~\cite{HowardJLP18, StekovicFL20, ZhangZZ20} tackle the room layout estimation beyond the cuboid shape assumption. \citet{HowardJLP18} leverage advanced detection methods~\cite{RenHGS15} to detect each plane instance and then reconstruct the layout from multiple posed images by a voting scheme. Built upon PlaneRCNN~\cite{LiuKGFK19}, \citet{StekovicFL20} formulate the layout reconstruction as constrained discrete optimization. However, this method requires several seconds to process a single frame, which is very time-consuming. \citet{ZhangZZ20} propose to regress the plane parameters and then utilize mean-shift clustering to get the plane segmentation. However, this method does not take the occlusion into account. All these methods merely consider plane cues to achieve reconstruction. In contrast, we jointly consider planes and lines, and show how the geometric relationship between them can assist the layout reconstruction.

Due to the limited field-of-view of the standard camera, another line of work~\cite{ZhangSTX14, ZouCSH18, SunHSC19, YangWPWSC19} proposes to exploit more contextual information from the panoramic images. For example, \citet{ZouCSH18} predict the corner maps and boundary maps directly. \citet{SunHSC19} propose to encode the layout as three 1D vectors, including ceiling-wall boundary, wall-wall boundary, and floor-wall boundary. \citet{YangWPWSC19} predict the floorplan probability in the ceiling view and floor view converted from the panorama.

\PAR{Single-view planar reconstruction.} Single-view piece-wise planar reconstruction~\cite{HoiemEH05} aims to use multiple planes to represent the scene. Most existing work~\cite{DelageLN07, MicusikWV08, BarinovaKYLLK08, ZaheerRK12} proposes to analyze the 2D geometric cues to recover 3D information, such as line segments, vanishing points. In contrast, recent CNN-based methods~\cite{LiuYCYF18, YangZ18, LiuKGFK19, YuZLZG19} have been proposed to tackle this problem in a top-down manner and achieve promising results. Compared to this problem, the room layout estimation is much more challenging due to strong occlusions by the foreground objects.

\begin{figure*}[t]
	\centering
	\includegraphics[width=0.85\linewidth]{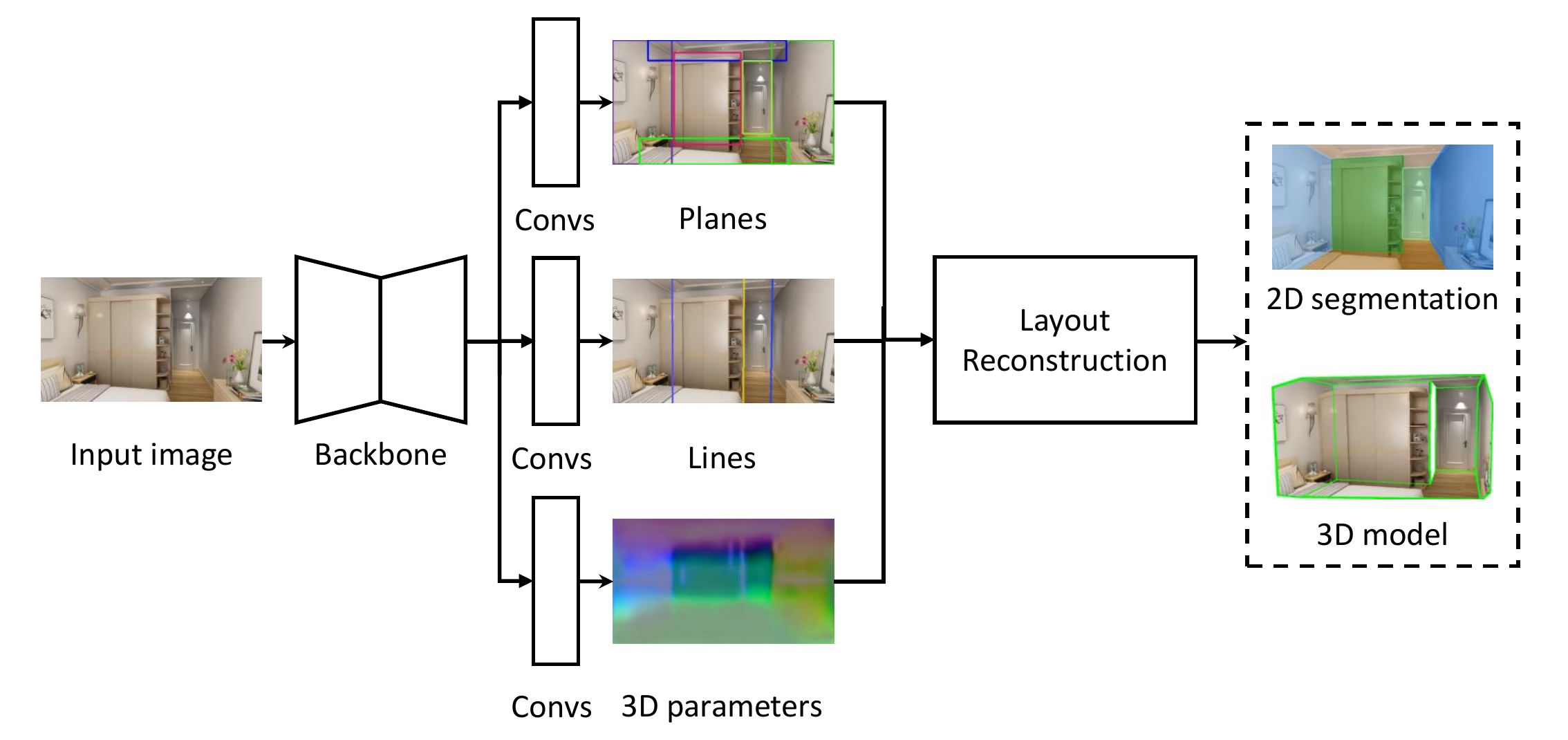}
	\caption{Pipeline. The network first takes a single RGB image as input and predicts planes and vertical lines. Meanwhile, we also estimate the 3D plane parameters for each plane. Then, a simple yet effective geometric reasoning method is adopted to reconstruct a geometrically consistent 3D room layout.}
	\label{fig:pipeline}
\end{figure*}

\PAR{Line segment and wireframe detection.} Line segment detection is a classical problem in computer vision. Conventional approaches to this problem involve grouping the low-level features in the image domain~\cite{Stephens91} or global accumulation in the Hough domain~\cite{GromponeJMR10}. Recently, \cite{LinPG20} trains deep networks with the Hough transformation priors to tackle this problem. Another line of work~\cite{HuangWZDGM18, ZhangLBZWHLXG19, ZhouQM19} proposes wireframe detection, \ie, jointly detecting straight line segments and how these lines connect to each other. In contrast, we only focus on a particular type of line segment, \ie, the vertical lines between adjacent wall planes.

\section{Our Method}

Our goal is to reconstruct the 3D room layout from a single RGB image. We first detect planes $\mathbf{P} = \{p_1, p_2, \ldots\}$ and vertical lines $\mathbf{L} = \{l_1, l_2, \ldots\}$. Meanwhile, we estimate the 3D parameters of the planes. Then, a simple yet effective geometric reasoning method is employed to reconstruct a geometrically consistent 3D room layout. \figref{fig:pipeline} shows the overall pipeline of our method.

\subsection{Plane and Line Detection}

Given an input RGB image $\mat{I} \in \mathbb{R}^{H \times W \times 3}$, we first adopt HRNet-W32~\cite{WangSCJDZLMTWLX20} as our backbone to extract the visual features:
\begin{equation}
	\mat{F} = \Call{Backbone}{\mat{I}},
\end{equation}
where $\mat{F} \in \mathbb{R}^{\hat{H} \times \hat{W} \times C}$. The resolution of the feature map is \num{4} times less than that of the input image, \ie, $H = 4 \hat{H}, W = 4 \hat{W}$.

Then, we use different CNN-based heads to detect the planes, vertical lines between adjacent walls and regress 3D parameters of planes, respectively.

\PAR{Planes.} Following CenterNet~\cite{ZhouWK19}, we represent each 2D plane $p_i$ using a bounding box including its center position $\vec{c}_i = (x_i, y_i)$ and size $\vec{s}_i = (w_i, h_i)$.

We use three branches to predict a plane center likelihood map $\mat{C} \in \mathbb{R}^{\hat{H} \times \hat{W} \times 3}$, a center offset map $\mat{O}^p \in \mathbb{R}^{\hat{H} \times \hat{W} \times 2}$, and a plane size map $\mat{S} \in \mathbb{R}^{\hat{H} \times \hat{W} \times 2}$. Each channel of the center likelihood map $\mat{C}$ represents semantic different categories, \ie, wall, floor, and ceiling. The corresponding ground truths are:
\begin{align}
	\mat{C} (\vec{u}) & = \exp \left(- \| \vec{u} - \lfloor \vec{c} \rfloor \|^2 / ( 2 {\delta^p}^2 ) \right), \\
	\mat{O}^p (\vec{u}) & =
	\begin{cases}
		\vec{c} - \vec{u}, & \vec{u} = \lfloor \vec{c} \rfloor, \\
		\mathbf{0}, & \textrm{otherwise},
	\end{cases} \\
	\mat{S} (\vec{u}) & =
	\begin{cases}
		\vec{s}, & \vec{u} = \lfloor \vec{c} \rfloor, \\
		\mathbf{0}, & \textrm{otherwise},
	\end{cases}
\end{align}
where $\vec{u} = (u_x, u_y)$ is the pixel coordinate on the output map and $\delta^p$ is an object size-adaptive standard deviation~\cite{LawD20}.

We use the focal loss~\cite{LinDGHHB17} to supervise the plane center likelihood map. For other maps, we use the standard $L_1$ loss and only calculate at the plane center locations.

\begin{figure*}
	\centering
	\setlength{\tabcolsep}{1pt}
	\begin{tabular}{cccc}
		\includegraphics[height=0.78in]{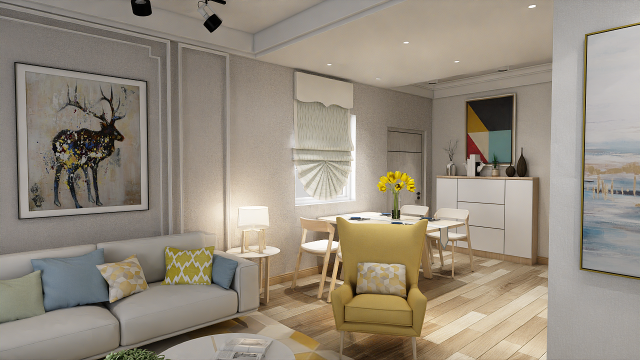} &
		\includegraphics[height=0.78in]{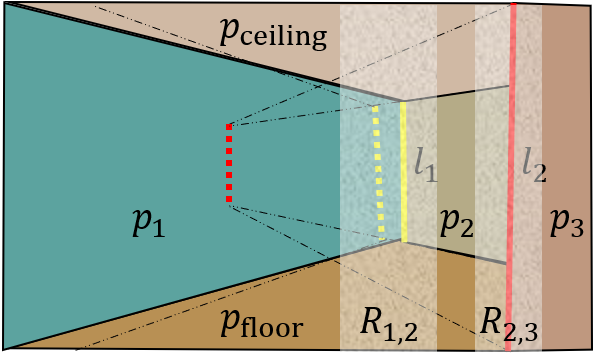} &
		\fbox{\includegraphics[height=0.78in]{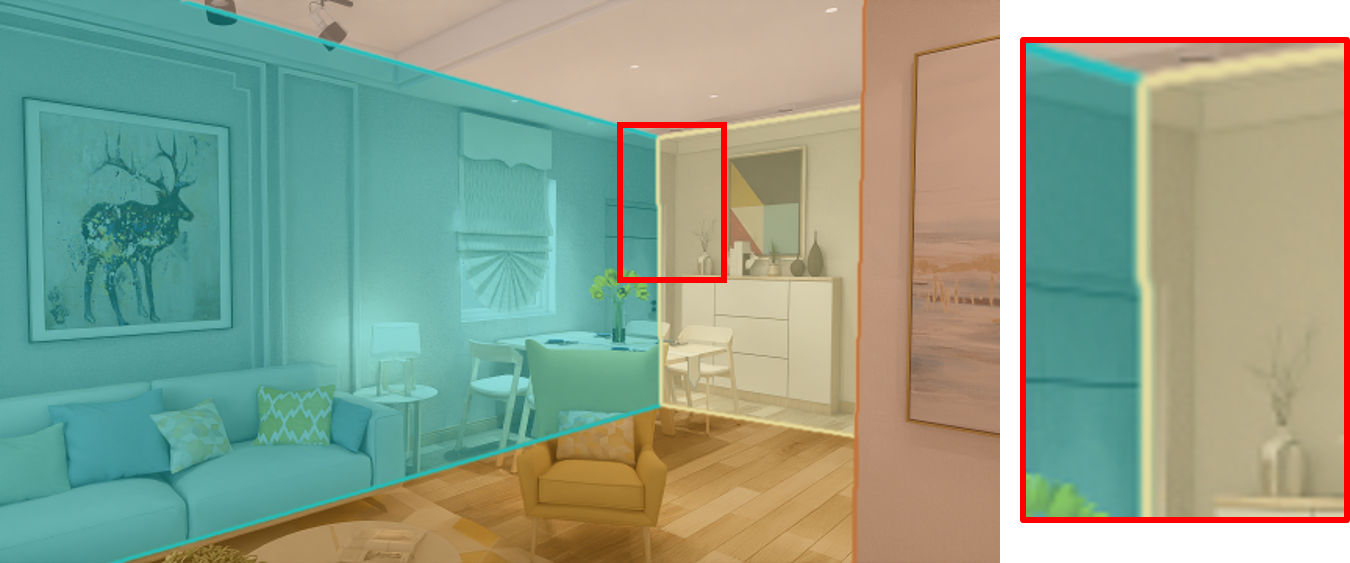}} &
		\fbox{\includegraphics[height=0.78in]{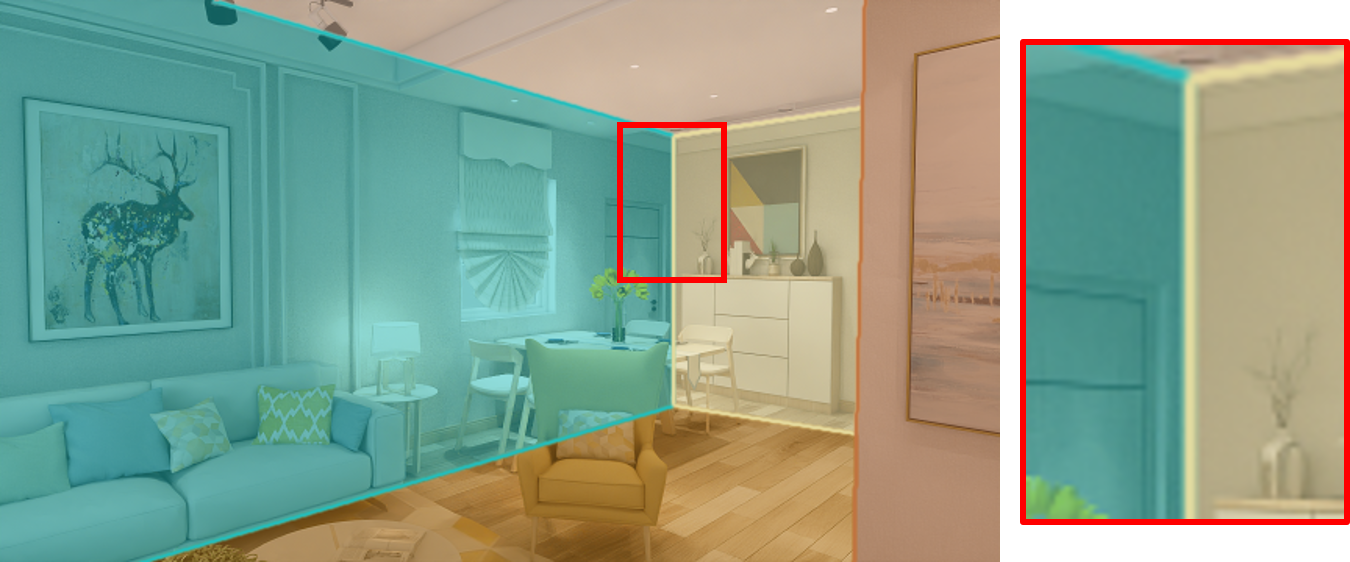}} \tabularnewline
		(a) & (b) & (c) & (d)
		\tabularnewline
	\end{tabular}
	\caption{Layout Reconstruction. (a) The input image. (b) Planes and lines candidates: planes (\ie, $p_1, p_2, p_3$, $p_\textrm{floor}$, and $p_\textrm{ceiling}$) and lines (\ie, the intersection line $l_1$ (solid yellow line), the occlusion line $l_2$ (solid red line)). The yellow dotted line is formed by the intersection of predicted $p_1$ and $p_2$. The red dotted line is formed by the intersection of predicted $p_2$ and $p_3$. $R_{1,2}$ and $R_{2,3}$ in the white shading represent the potential intersection line region of $p_1$ and $p_2$, $p_2$ and $p_3$, respectively. (c) 2D layout segmentation before optimization. (d) 2D layout segmentation after optimization.}
	\label{fig:optimize}
\end{figure*}

\PAR{Lines.} For a 2D vertical line $l_j$ between two adjacent walls, we represent it by its angle $\theta_j$ of the line inclination and points $\mathcal{T}_j$ lying on the line:
\begin{multline}
	\mathcal{T}_j = \{ \vec{t}_i = (t_{i,x}, t_{i,y}) \mid t_{i,y} \in [ y_\textrm{min}, y_\textrm{max} ], \\ t_{i,y} \in \mathbb{N}, t_{i,x} \in \mathbb{R} \},
\end{multline}
where $y_\textrm{min}$ and $y_\textrm{max}$ represent the upper and lower bounds along the $y$-axis in the output map, respectively.

We adopt another three branches to predict a line likelihood map $\mat{L} \in \mathbb{R}^{\hat{H} \times \hat{W}}$, an offset map $\mat{O}^l \in \mathbb{R}^{\hat{H} \times \hat{W}}$, and an orientation map $\mat{\Theta}\in \mathbb{R}^{\hat{H} \times \hat{W}}$:
\begin{align}
	\mat{L} (\vec{u}) & =
	\begin{cases}
		\exp \left( - {\| \vec{u} - \lfloor \vec{t}_i \rfloor \|^2} / (2 {\delta^l}^2) \right), & u_y = t_{i,y}, \\
		0, & \textrm{otherwise},
	\end{cases} \\
	\mat{O}^l(\vec{u}) & =
	\begin{cases}
		t_{i,x} - u_x, & \vec{u} = \lfloor \vec{t}_i \rfloor, \\
		0, & \textrm{otherwise},
	\end{cases} \\
	\mat{\Theta} (\vec{u}) & =
	\begin{cases}
		\theta_j, & \vec{u} = \lfloor \vec{t}_i \rfloor, \\
		0, & \textrm{otherwise},
	\end{cases}
\end{align}
where $\delta^l=5/6$. Specifically, the offset along $y$-axis is always \num{0}, so we only predict the offset along the $x$-axis.

We use the focal loss~\cite{LinDGHHB17} to supervise the line region likelihood map. For other maps, we use the standard $L_1$ loss and only calculate at the line locations.

\PAR{3D plane parameters.} To reconstruct the 3D room layout, we further estimate the 3D parameters for each plane. The 3D plane parameters include its surface normal $\vec{n} \in \mathbb{S}^2$ and offset $d$. For a 3D point $\vec{q} \in \mathbb{R}^3$ lying on the plane, we have $\vec{n}^T \vec{q} + d = 0$. Let $\vec{v} = [ \vec{n}, d ]$, we predict a plane parameter map $\mat{V} \in \mathbb{R}^{\hat{H} \times \hat{W} \times 3}$ at the plane centers:
\begin{align}
	\mat{V} (\vec{u}) & =
	\begin{cases}
		\vec{v}, & \vec{u} = \lfloor \vec{c} \rfloor, \\
		0, & \textrm{otherwise}.
	\end{cases}
\end{align}

We use the standard smooth $L_1$ loss~\cite{Girshick15} to supervise the learning of the plane parameters. Inspired by PlaneRecover~\cite{YangZ18} and PlaneAE~\cite{YuZLZG19}, we also use the standard smooth $L_1$ loss to supervise the depth map inferred by plane parameters.

\PAR{Inference.} During inference, we extract plane and line candidates from the outputs. Following CenterNet~\cite{ZhouWK19}, we first find all peaks $(x_i, y_i)$ in the likelihood map for each semantic category of the planes. The corresponding offset, bounding box size and 3D plane parameter can be obtained:
\begin{align}
	\{o_{i,x}, o_{i,y}\} & = \mat{O}^p (x_i, y_i), \\
	\{w_i, h_i\} & = \mat{S} (x_i, y_i), \\
	\vec{v} & = \mat{V} (x_i, y_i).
\end{align}
Then, the bounding box of the detected plane is $(x_i + o_{i,x}, y_i + o_{i,y}, w_i, h_i)$.

For lines, we use the normal form of the line, \ie, $x \cos \theta + y \sin \theta - b = 0$. Similarly, we first extract the peaks $(x_i, y_i)$ in the likelihood map. The corresponding offset and line inclination can be obtained:
\begin{align}
	o_{i,x} & = \mat{O}^l (x_i, y_i), \\
	\theta & = \mat{\Theta} (x_i, y_i).
\end{align}
Then, we obtain $b = (x_i + o_{i,x}) \cos \theta + y_i \sin \theta$.

We use the non-maximum suppression (NMS) to remove the duplicated candidates. For planes, we use IoU-based NMS among all categories. For lines, if they intersect in the image or the maximal distance along $x$ coordinate with each row is less than a threshold, we discard the one with the lower confidence.

\subsection{3D Layout Reconstruction}

Once we have detected planes, vertical lines, and 3D parameters of planes, we further perform geometric reasoning to reconstruct the 3D room layout under the assumption that the room layout consists of a single ceiling, a single floor, and several vertical walls. In such an assumption, any two non-adjacent walls in image space must be physically disconnected in 3D space, so we only need to infer the connectivity relations of the adjacent walls. Once we know the connectivity relations of all walls, we combine the floor and the ceiling to achieve reconstruction.

Specifically, the detected planes contain several vertical walls, a ceiling, and a floor. We first order all walls from the lowest to the highest by the $x$-coordinates of plane centers in image space. Next, we calculate the intersection line of two adjacent walls by their 3D parameters and project it into image space with the known camera intrinsic matrix. Then, we classify the geometric relationship in 3D space of these two walls into two types: physically connected or disconnected, depending on whether the projected line lies between two adjacent walls or not. More specifically, we define a potential intersection line region by the bounding boxes of the adjacent walls in image space, as shown in \figref{fig:optimize}(b). Each potential intersection line region has at most one detected line, and we choose the one with the highest confidence when there are multiple detected lines. Due to space limitations, we refer readers to supplementary material for more details.

If two walls are physically connected in 3D space, we directly calculate the boundary with their 3D parameters. Furthermore, we expect to construct a geometrically consistent 3D room layout between detected planes and lines. We optimize the 3D plane parameters to align the calculated boundary with the detected intersection line. Specifically, we construct a list of triplets $\mathcal{T} = \{ (p_i, l_j, p_k) \}$ by identifying all detected intersection lines $l_j$, and the wall planes $p_i, p_k$ on its two sides. Then, we optimize the predicted 3D plane parameters $(\vec{n}, d)$ by the following objective function:
\begin{equation}
	\begin{split}
		\min_{\vec{n}, d} \sum_{(i, j, k) \in \mathcal{T}} & \| ( \frac{\vec{n}_i}{d_i} - \frac{\vec{n}_k}{d_k} )^T \mat{K}^{-1} - \vec{t}_j^T \| \\
		+ & \alpha \| \vec{n}_i - \tilde{\vec{n}}_i \| + \beta \| d_i - \tilde{d}_i \| \\
		+ & \alpha \| \vec{n}_k - \tilde{\vec{n}}_k \| + \beta \| d_k - \tilde{d}_k \|,
		\label{eq:optimize}
	\end{split}
\end{equation}
where $\vec{t}_j = [\cos\theta, \sin\theta, -b]^T$ is the detected intersection line parameter in the homogeneous coordinate, $\mat{K} \in \mathbb{R}^{3 \times 3}$ is the known camera intrinsic matrix, $\tilde{\vec{n}}$ and $\tilde{d}$ are the plane parameters estimated by the neural network, $\alpha$ and $\beta$ are balance weights. In our experiment, we set $\alpha = 1$ and $\beta = 0.01$. The first term enforces the two connected wall planes to fit the detected line, and the rest keeps the solution close to the initial estimated plane parameters. We use L-BFGS optimization method~\cite{NocedalW06} from the SciPy library to solve this problem. Figures~\ref{fig:optimize}(c) and (d) show the qualitative comparison with or without the proposed layout optimization. As expected, the results with optimization preserve the boundary of the walls very well.

In contrast, if two walls are physically disconnected, and the occlusion occurs. Instead of fitting the occlusion line to the points with the largest depth discrepancy changes~\cite{StekovicFL20}, we directly use the detected line as the occlusion line to handle the occlusion problem. Specifically, when a detected line is located between two disconnected walls, we regard it as the occlusion line. However, when failing to detect the occlusion line, we coarsely locate it by the bounding boxes of wall planes. Then, same as RaC~\cite{StekovicFL20}, we also introduce a virtual plane $p$, \ie, back-projection of the occlusion line, defined by the camera center and the occlusion line. Since the virtual plane passes through the camera center, we obtain the offset $d = 0$. The surface normal $\vec{n}$ of the virtual plane can be obtained as:
\begin{align}
	\vec{n} = \frac{\mat{K}^T \vec{t}}{\| \mat{K}^T \vec{t} \|}.
	\label{eq:virtual}
\end{align}

By introducing the virtual plane passing through the occlusion line, all adjacent walls in the image space are physically connected in the 3D space. Thus, we directly calculate the boundary of adjacent walls with their 3D parameters. Specifically, we calculate the 3D corner $\vec{q} \in \mathbb{R}^3$ of the room layout by solving a system of linear equations:
\begin{align}
	\left\lbrace \begin{array}{l}
		\vec{n}_i^T \vec{q} + d_i = 0, \\
		\vec{n}_j^T \vec{q} + d_j = 0, \\
		\vec{n}_k^T \vec{q} + d_k = 0, \\
	\end{array}	\right.
	\label{eq:3dpoint}
\end{align}
where $i, j$ are indices of the two adjacent wall planes, and $k$ represents either a floor or a ceiling plane. %

Then, we obtain the 2D corner $\vec{p}$ by projecting the 3D point $\vec{q}$ to the 2D image space:
\begin{align}
	\vec{p} \sim \mat{K} \vec{q}.
	\label{eq:2dpoint}
\end{align}

\section{Experiments}

In this section, we conduct experiments to evaluate the performance of the proposed method over two public benchmarks: Structured3D dataset~\cite{ZhengZLTGZ20} and NYUv2 303 dataset~\cite{SilbermanHKF12, ZhangKSU13}.

\subsection{Implementation Details}

We implement our network with PyTorch~\cite{PyTorch}. The batch size is set to \num{24}. We use color jittering as data augmentation. For Structure3D dataset, we train the model for \num{50} epochs. We use Adam optimizer~\cite{KingmaB15} with learning rate \num{1e-4} and \num{5e-4} weight decay. The learning rate is decayed by a factor of \num{10} at \num{30}\textsuperscript{th} and \num{40}\textsuperscript{th} epoch. For NYUv2 303 dataset, we adopt the model trained on Structure3D dataset and fine-tune it on the SUN RGB-D dataset~\cite{SongLX15} for \num{50} epochs. Then, we fine-tune the model on NYUv2 subset for \num{10} epochs. The learning rate is set to \num{1e-4}. Since NYUv2 303 dataset is a subset of the SUN RGB-D dataset, we exclude \num{101} test images from the training set.

\subsection{Results on Structured3D Dataset}

Structured3D dataset~\cite{ZhengZLTGZ20} is a large-scale photo-realistic synthetic dataset with ground-truth 3D room structure annotations. We divide the dataset at the scene level into train/validation/test, which contain \num{3000}/\num{250}/\num{250} scenes and \num{68096}/\num{6579}/\num{6280} images. The resolution of input image is $640 \times 384$.

\begin{figure*}[t]
	\centering
	\setlength{\tabcolsep}{1pt}
	\begin{tabular}{ccccc}
		\includegraphics[width=0.18\linewidth]{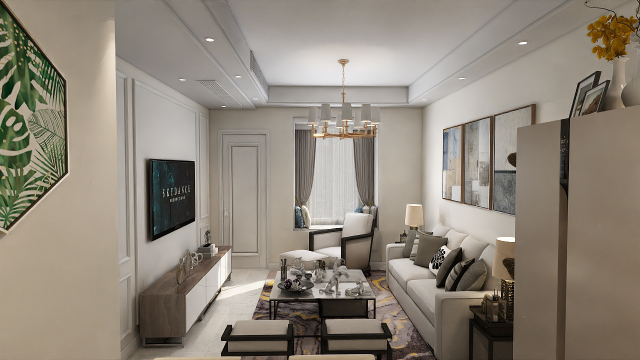} &
		\includegraphics[width=0.18\linewidth]{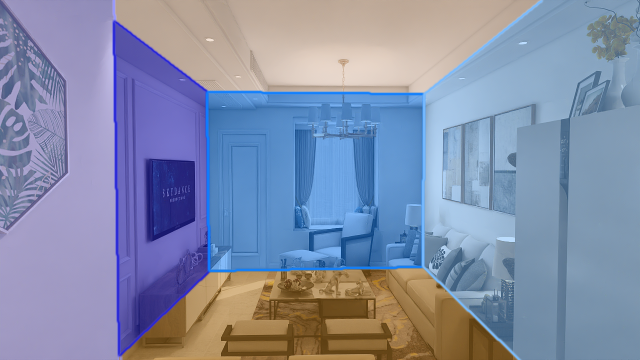} &
		\includegraphics[width=0.18\linewidth]{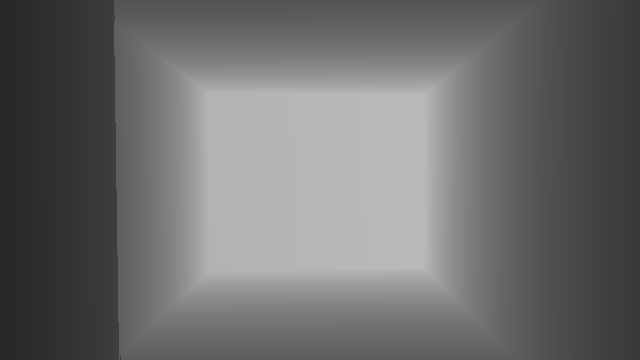} &
		\includegraphics[width=0.18\linewidth]{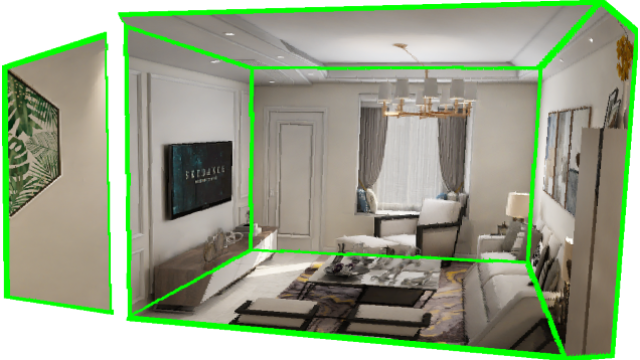} &
		\includegraphics[width=0.18\linewidth]{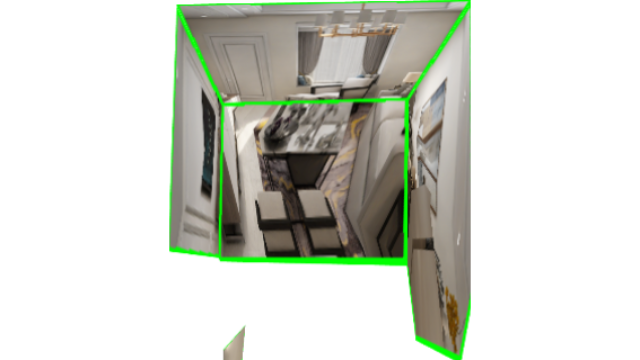}
		\tabularnewline
		\includegraphics[width=0.18\linewidth]{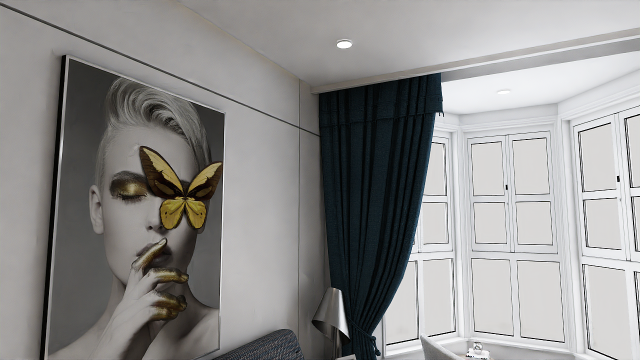} &
		\includegraphics[width=0.18\linewidth]{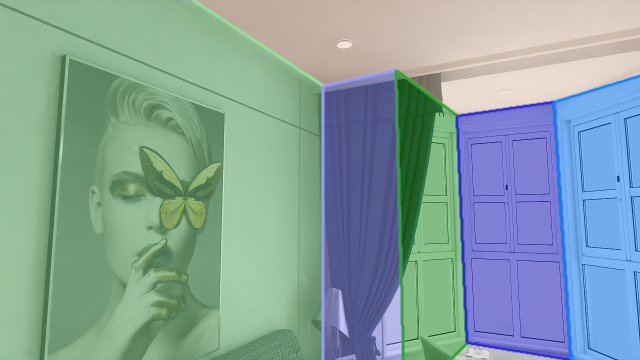} &
		\includegraphics[width=0.18\linewidth]{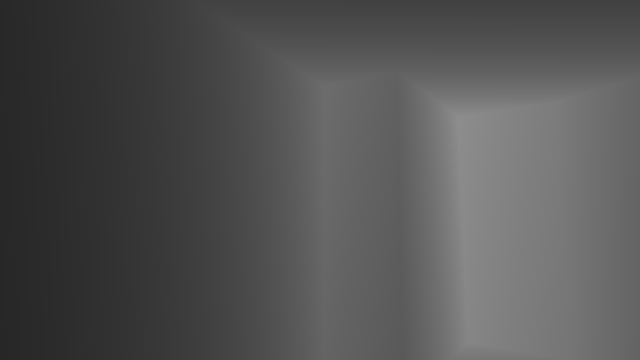} &
		\includegraphics[width=0.18\linewidth]{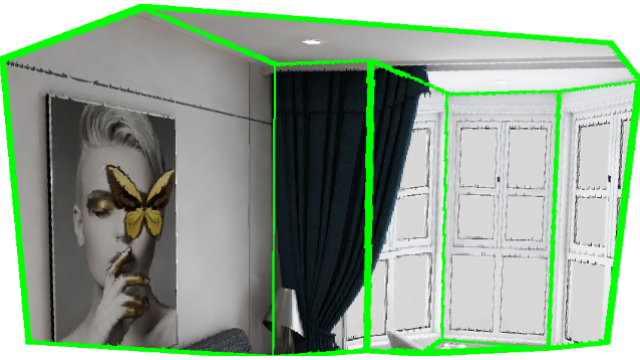} &
		\includegraphics[width=0.18\linewidth]{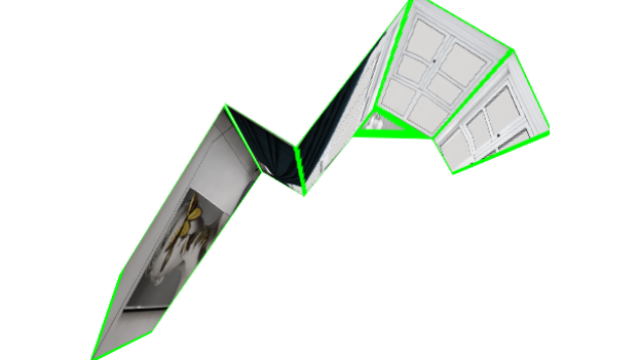}
		\tabularnewline
		\includegraphics[width=0.18\linewidth]{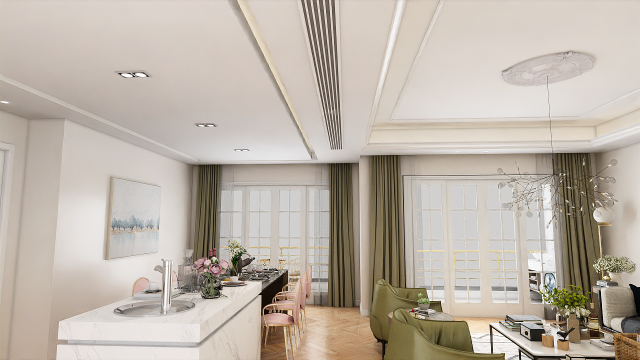} &
		\includegraphics[width=0.18\linewidth]{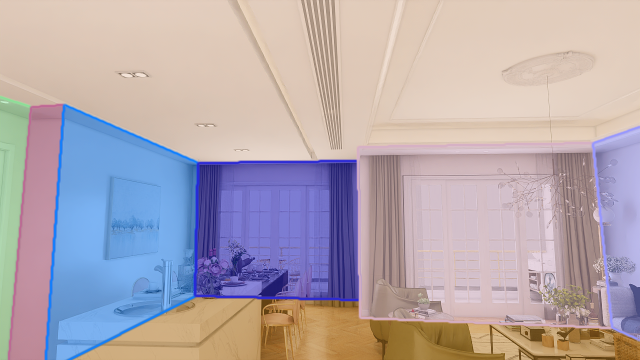} &
		\includegraphics[width=0.18\linewidth]{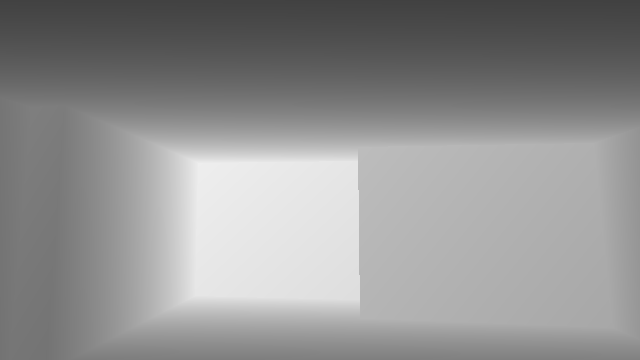} &
		\includegraphics[width=0.18\linewidth]{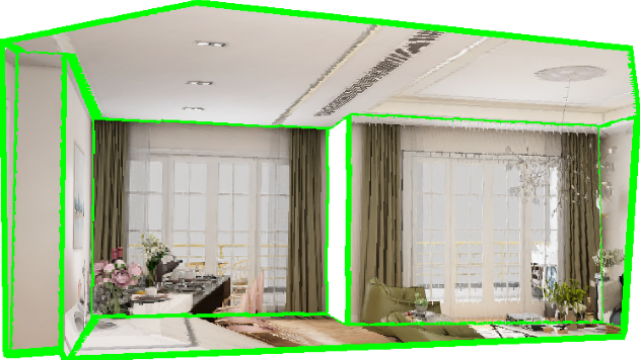} &
		\includegraphics[width=0.18\linewidth]{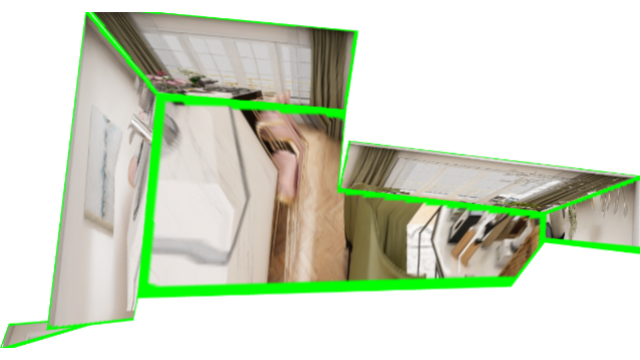}
		\tabularnewline
		\includegraphics[width=0.18\linewidth]{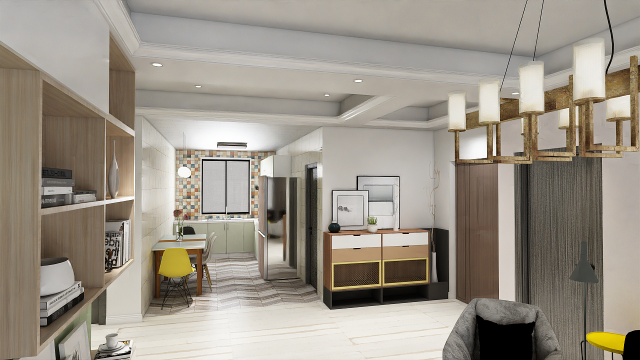} &
		\includegraphics[width=0.18\linewidth]{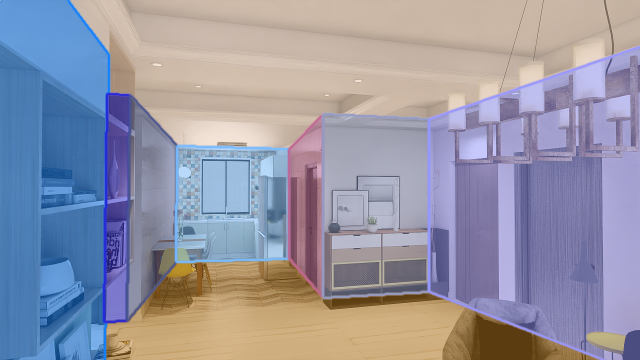} &
		\includegraphics[width=0.18\linewidth]{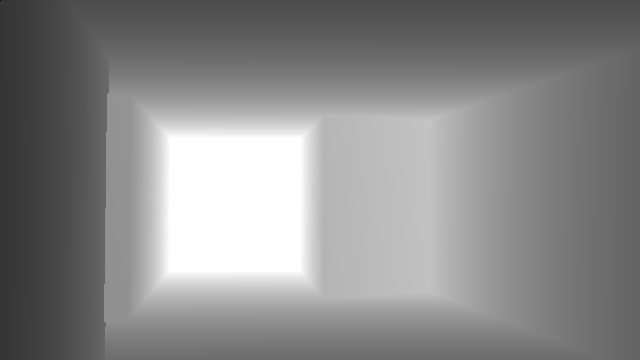} &
		\includegraphics[width=0.18\linewidth]{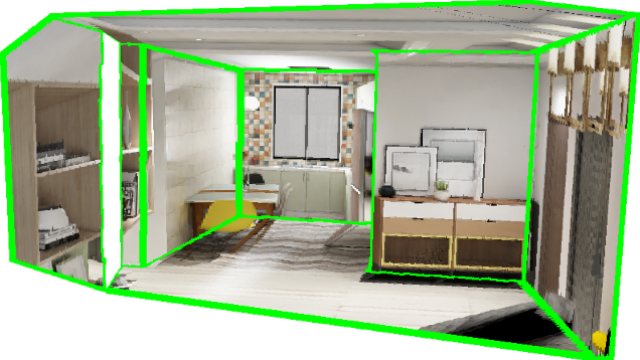} &
		\includegraphics[width=0.18\linewidth]{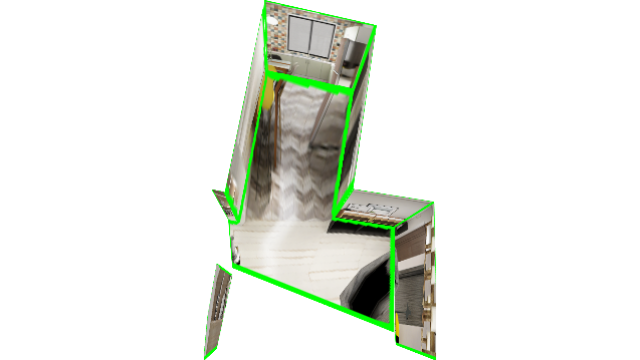}
		\tabularnewline
		\includegraphics[width=0.18\linewidth]{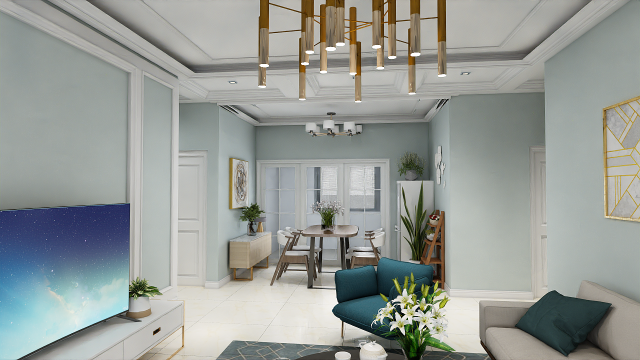} &
		\includegraphics[width=0.18\linewidth]{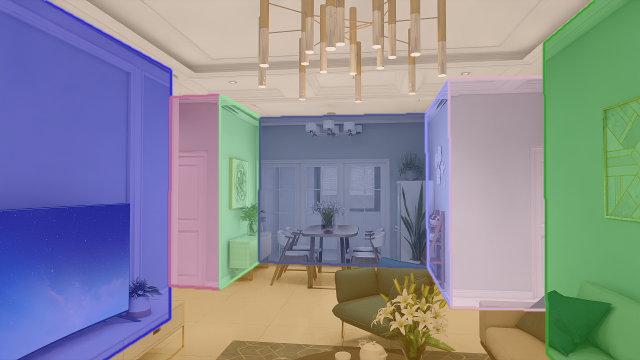} &
		\includegraphics[width=0.18\linewidth]{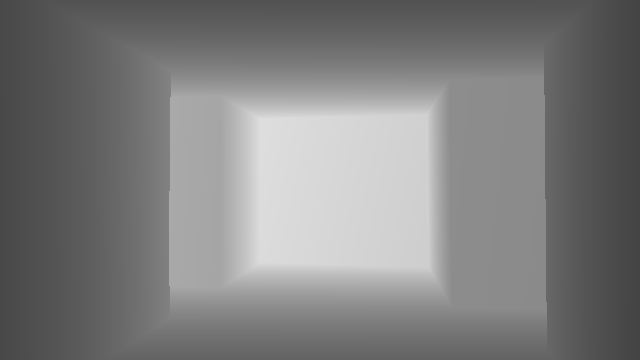} &
		\includegraphics[width=0.18\linewidth]{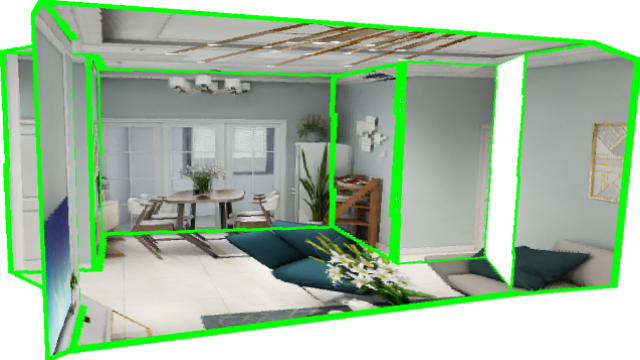} &
		\includegraphics[width=0.18\linewidth]{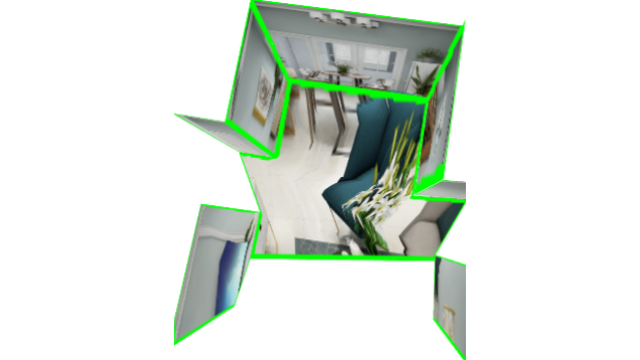}
		\tabularnewline
		Input image & 2D layout & Depth map & 3D model & 3D model (top-view)
		\tabularnewline
	\end{tabular}
	\caption{3D room layout reconstruction results on Structured3D dataset~\cite{ZhengZLTGZ20}. The ceiling is ignored in the top view of the 3D model. More reconstruction results can be found in the supplementary material.}
	\label{fig:s3d:3d}
\end{figure*}

\begin{table*}[t]
	\centering
	\sisetup{
		table-number-alignment 	= center,
		table-figures-integer 	= 2,
		table-figures-decimal 	= 2,
		table-auto-round 		= true,
		detect-weight			= true,
	}
	\begin{tabular}{lSSSS[table-figures-decimal=4]S}
		\toprule
		{Method} & {IoU (\%) $\uparrow$} & {PE (\%) $\downarrow$} & {EE $\downarrow$} & {RMSE $\downarrow$} & {Runtime (\si{\second}) $\downarrow$}
		\tabularnewline
		\midrule
		{Planar R-CNN~\cite{HowardJLP18}$^\dagger$} & 79.6380106 & 7.03876450 & 6.57873658 & 0.401323167 & 0.11229391 \tabularnewline
		\midrule
		{RaC~\cite{StekovicFL20}} & 76.2910162 & 8.07076333 & 7.18793163 & 0.346485293 & 5.3523912 \tabularnewline
		{Ours (w/o optimization)} & 79.9414766 & 6.40490779 & 6.79561008 & \bfseries 0.282730500 & \bfseries 0.07482198 \tabularnewline
		{Ours} & \bfseries 81.3968758 & \bfseries 5.87224901 & \bfseries 5.78071385 & 0.290529978 & 0.24317952 \tabularnewline
		\bottomrule
	\end{tabular}
	\caption{Quantitative results on Structured3D dataset~\cite{ZhengZLTGZ20}. $\dagger$: our implementation.}
	\label{tab:s3d}
\end{table*}

\PAR{Methods for comparison.} We compare our method with the following two methods: (i) Planar R-CNN~\cite{HowardJLP18}: Although Planar R-CNN can reconstruct the piece-wise planar model from a single image, it does not reason about the extents of the planar surface, which is vital in the room layout estimation task. Since the source code of Planar R-CNN is unavailable, we reimplement it closely following the given implementation details. (ii) RaC~\cite{StekovicFL20}\footnote{\url{https://github.com/vevenom/RoomLayout3D_RandC}}: We use the plane detection results by our methods as the input. Additionally, this method requires plane instance segmentation. To this end, we further predict the plane parameters for each pixel. During inference, we compare the pixel-level plane parameters with the instance-level plane parameters to get the plane instance segmentation. Since the above two methods build on 3D plane detection results, and the key contribution is the different post-process steps to reconstruct the room layout according to the plane detection results, for a fair comparison, we use the same plane detection results as our method.

\begin{figure*}[t]
	\centering
	\setlength{\tabcolsep}{1pt}
	\begin{tabular}{cccccc}
		\includegraphics[width=0.15\linewidth]{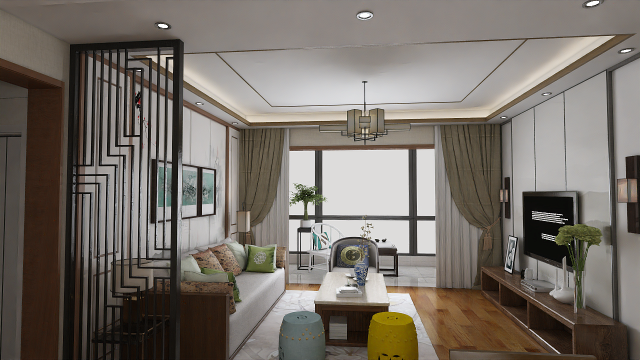} &
		\includegraphics[width=0.15\linewidth]{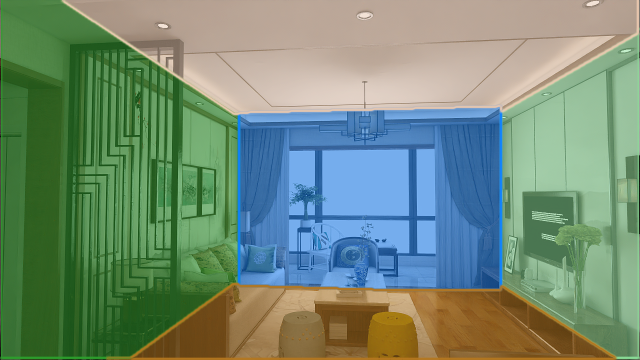} &
		\includegraphics[width=0.15\linewidth]{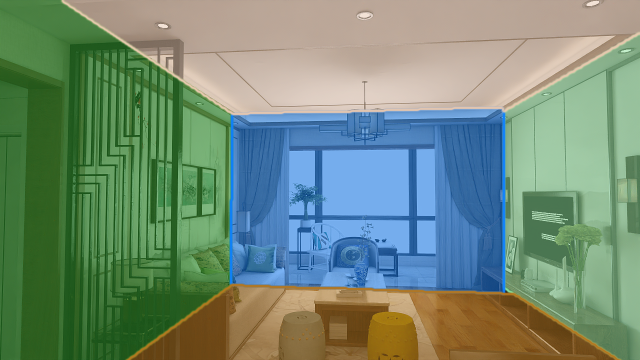} &
		\includegraphics[width=0.15\linewidth]{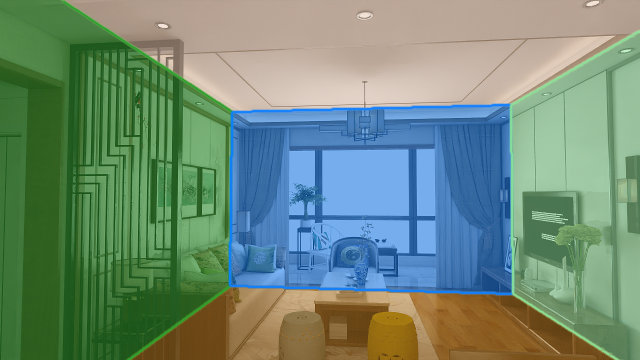} &
		\includegraphics[width=0.15\linewidth]{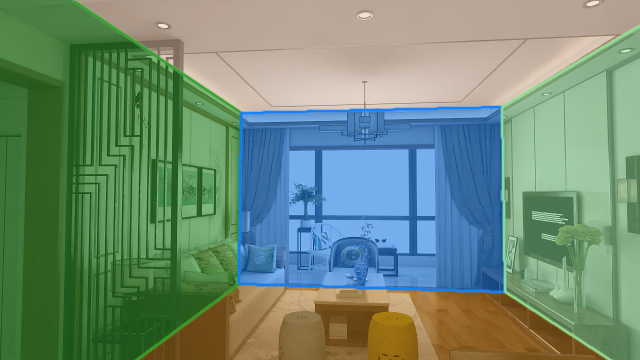} &
		\includegraphics[width=0.15\linewidth]{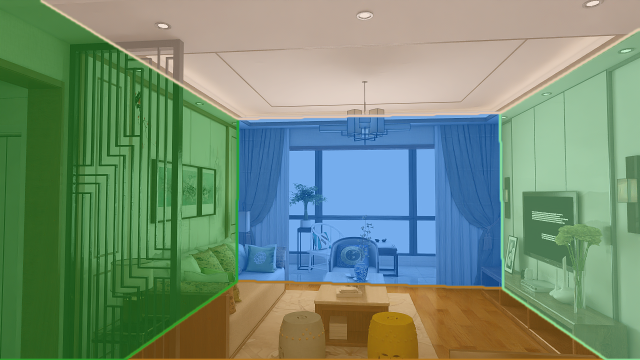}
		\tabularnewline
		\includegraphics[width=0.15\linewidth]{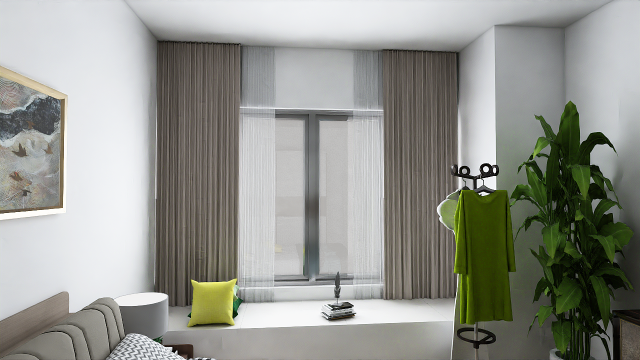} &
		\includegraphics[width=0.15\linewidth]{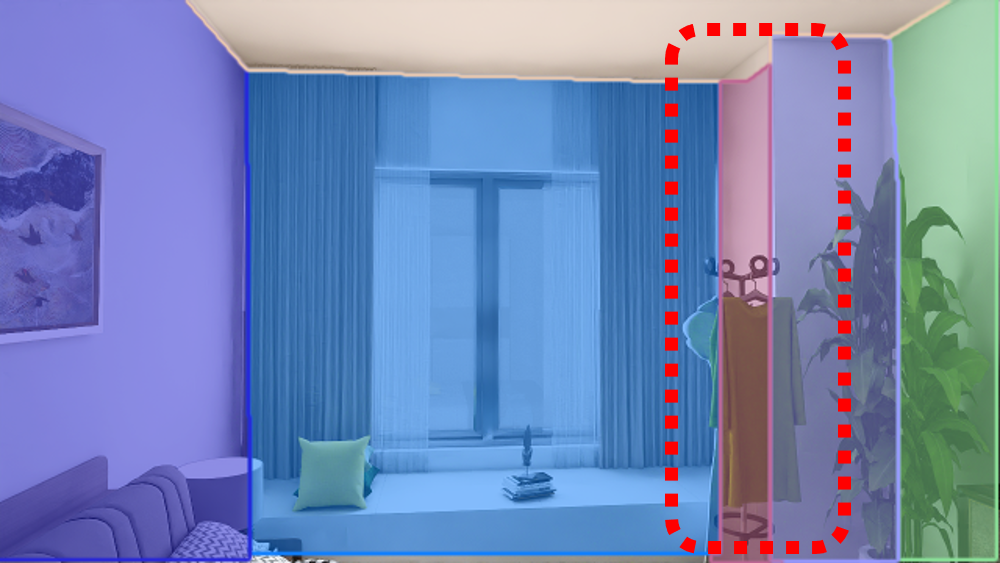} &
		\includegraphics[width=0.15\linewidth]{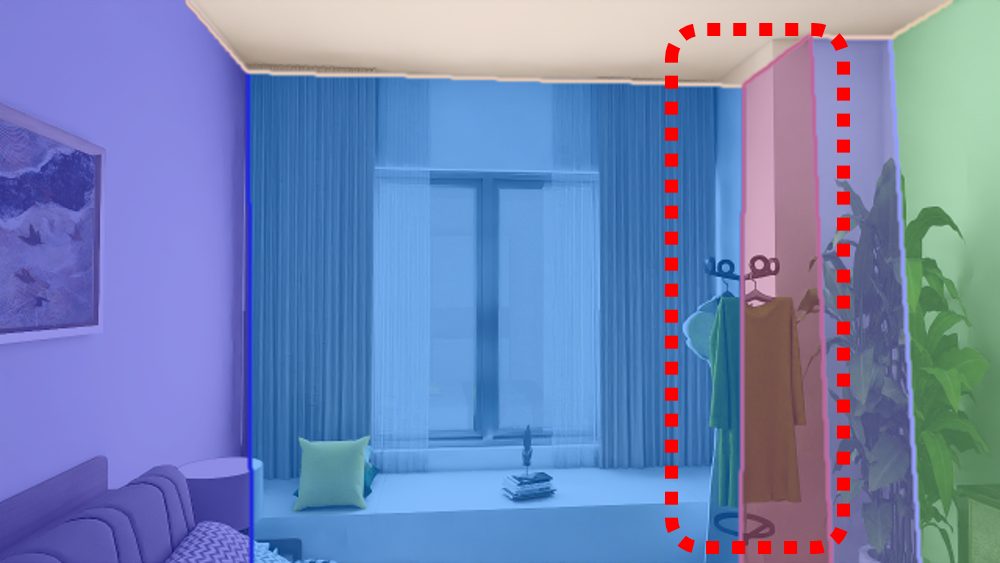} &
		\includegraphics[width=0.15\linewidth]{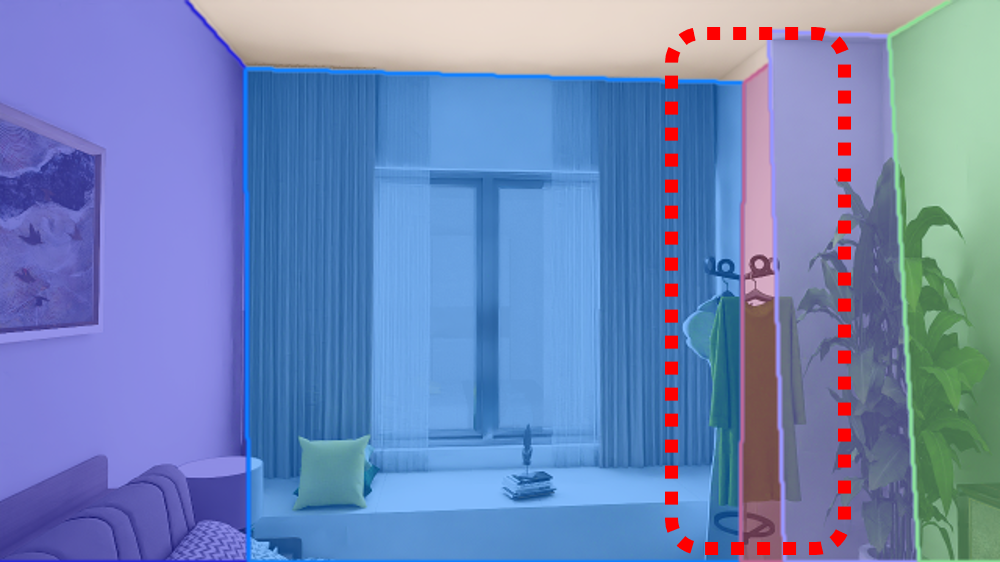} &
		\includegraphics[width=0.15\linewidth]{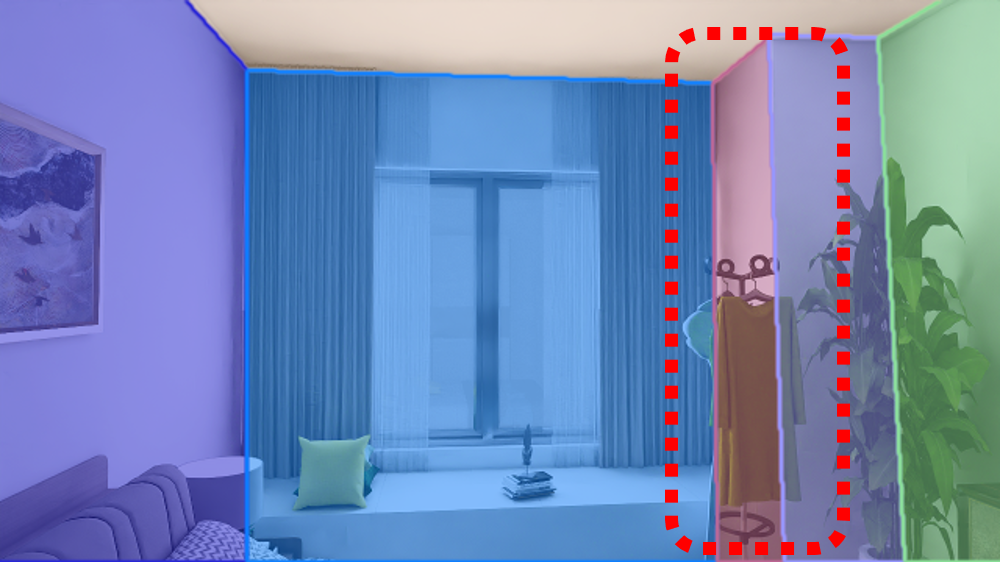} &
		\includegraphics[width=0.15\linewidth]{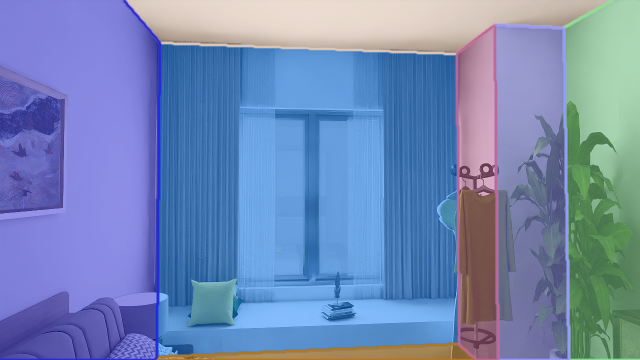}
		\tabularnewline
		\includegraphics[width=0.15\linewidth]{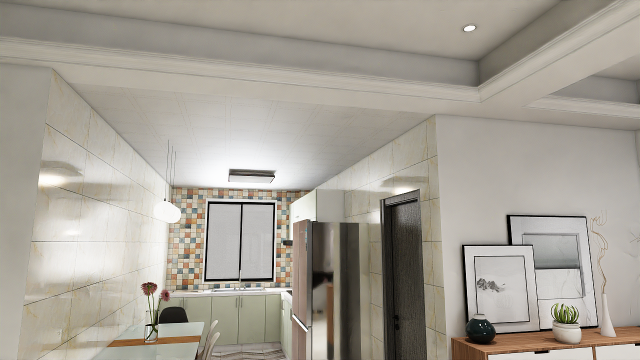} &
		\includegraphics[width=0.15\linewidth]{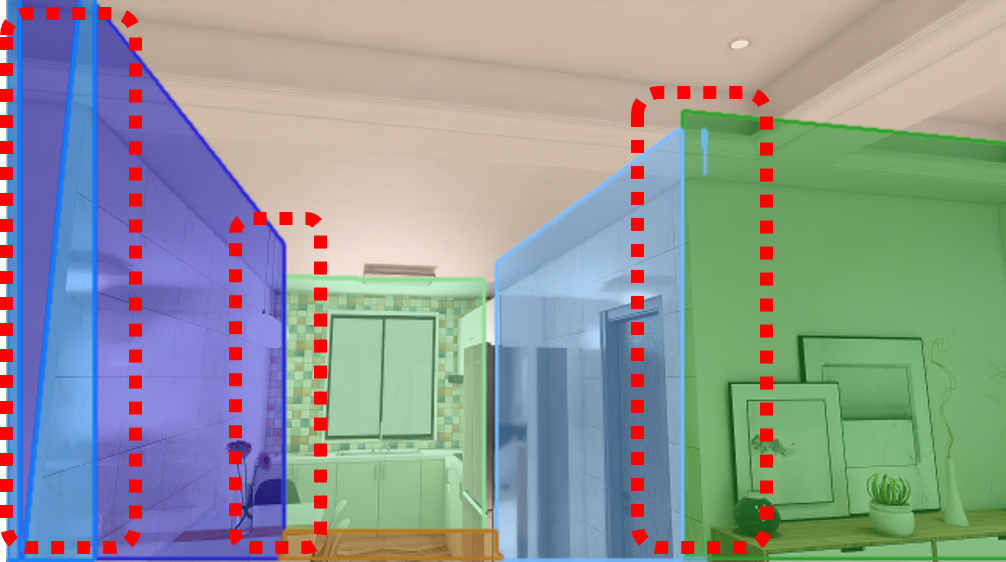} &
		\includegraphics[width=0.15\linewidth]{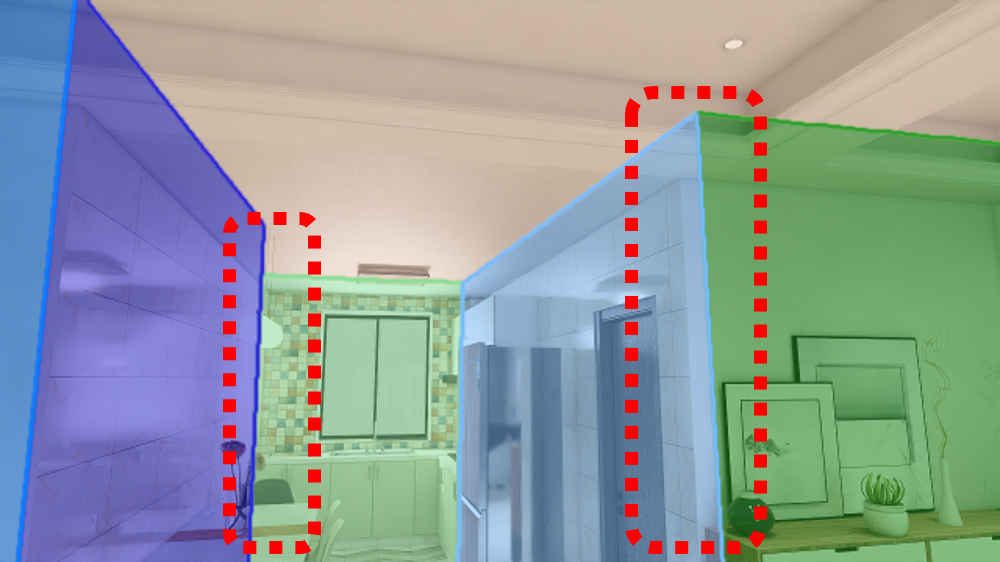} &
		\includegraphics[width=0.15\linewidth]{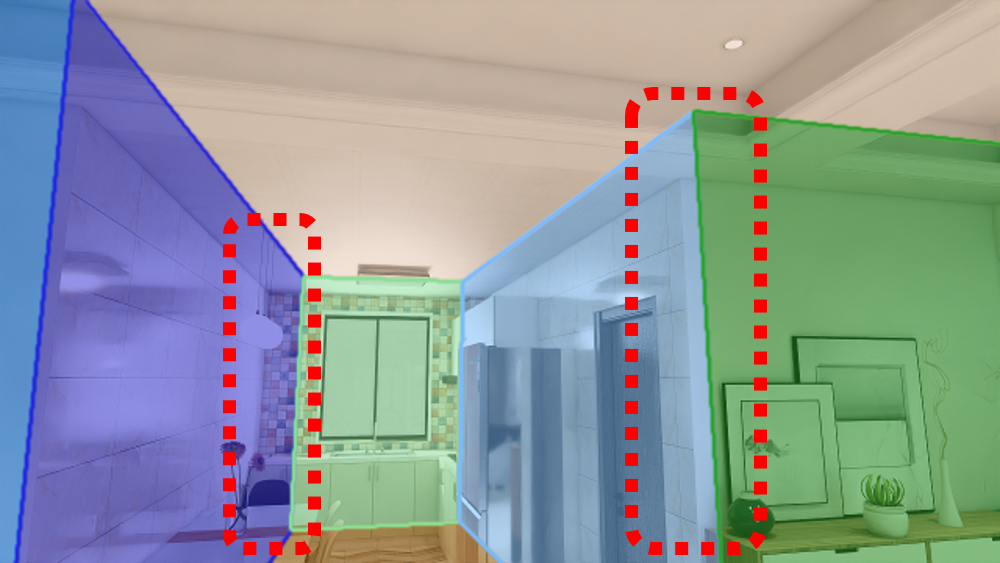} &
		\includegraphics[width=0.15\linewidth]{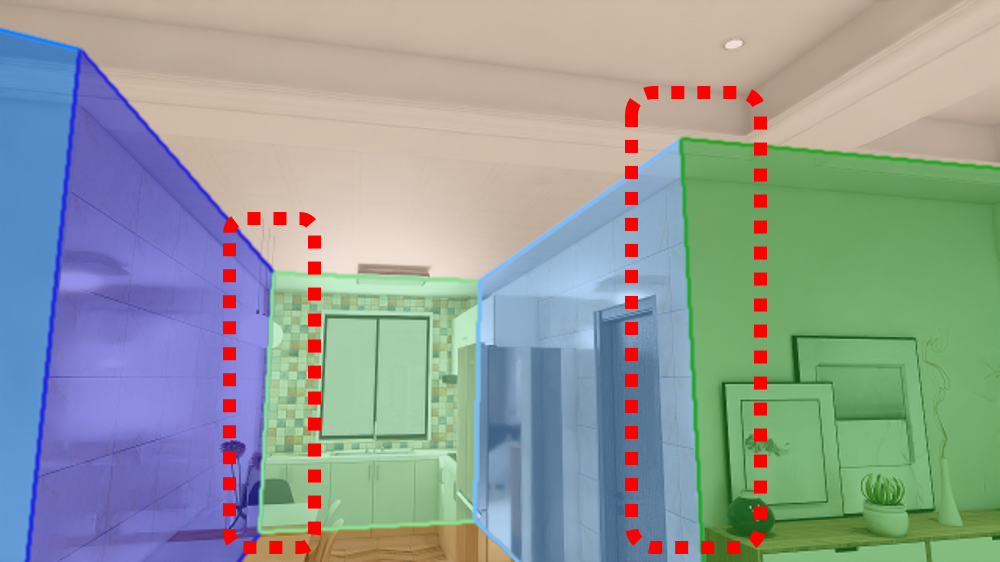} &
		\includegraphics[width=0.15\linewidth]{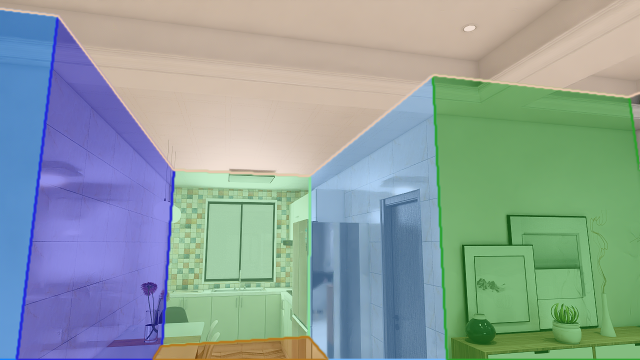}
		\tabularnewline
		\includegraphics[width=0.15\linewidth]{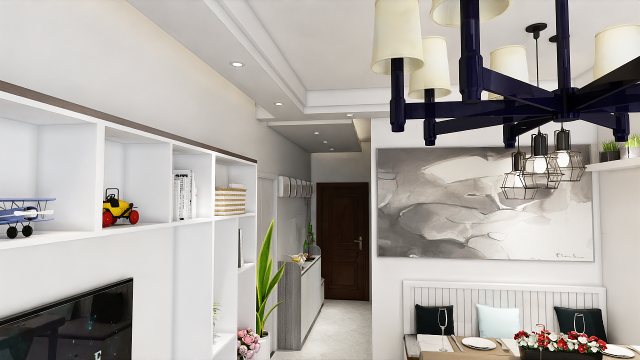} &
		\includegraphics[width=0.15\linewidth]{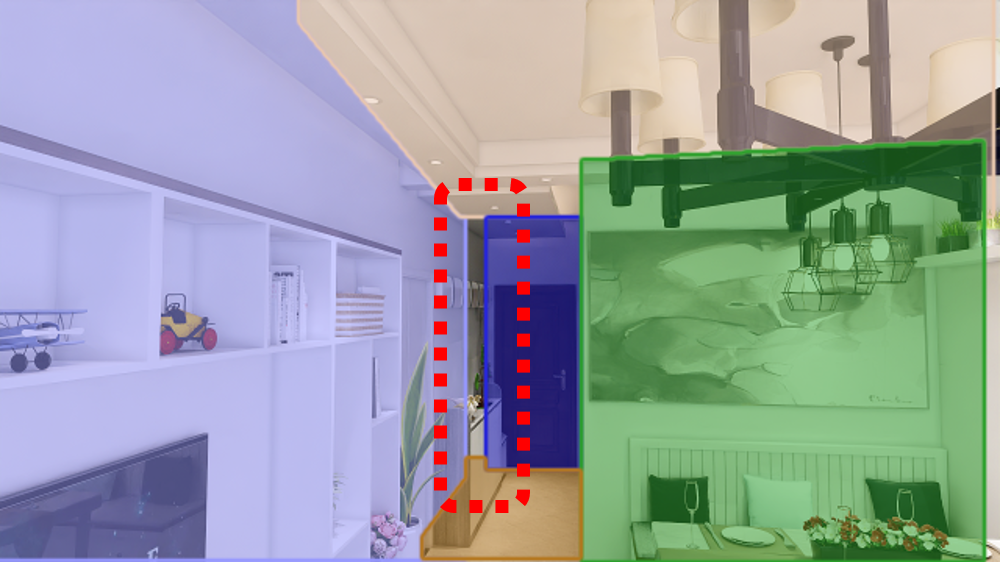} &
		\includegraphics[width=0.15\linewidth]{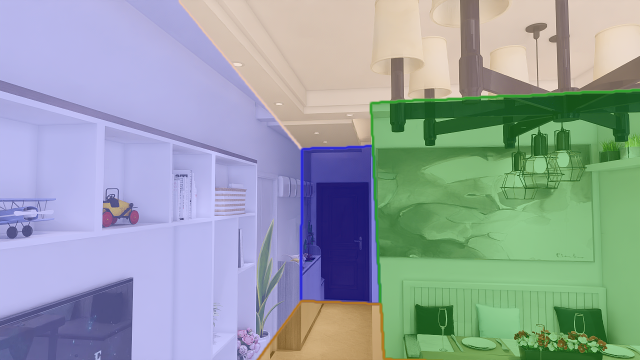} &
		\includegraphics[width=0.15\linewidth]{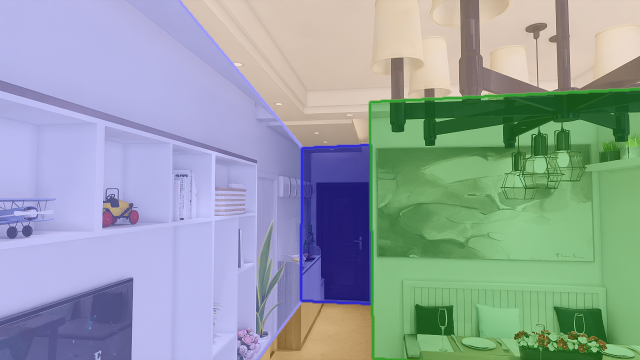} &
		\includegraphics[width=0.15\linewidth]{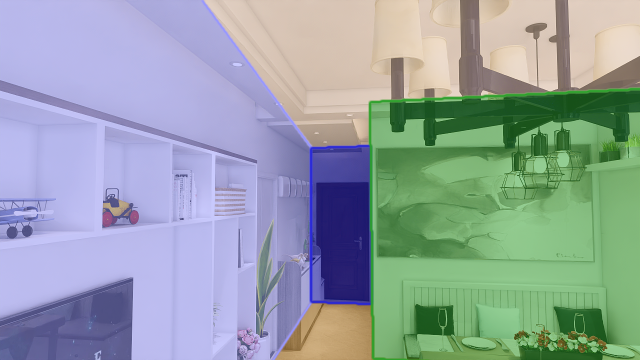} &
		\includegraphics[width=0.15\linewidth]{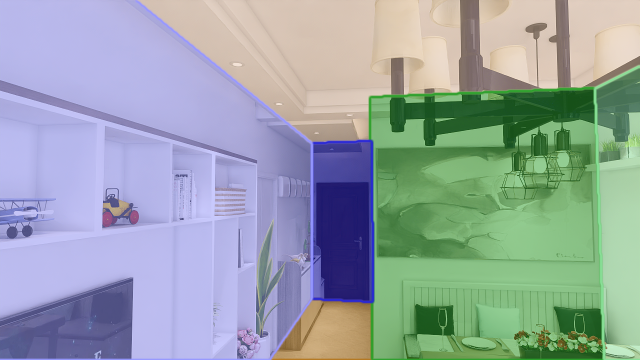}
		\tabularnewline
		\includegraphics[width=0.15\linewidth]{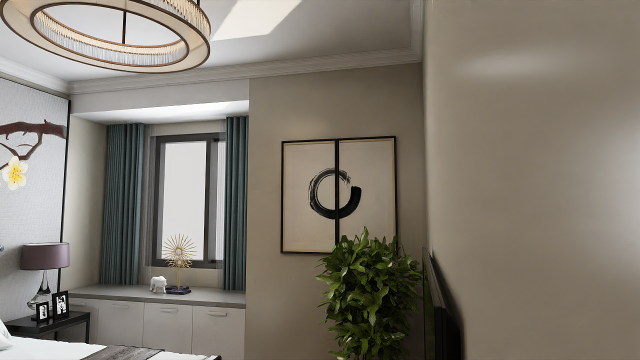} &
		\includegraphics[width=0.15\linewidth]{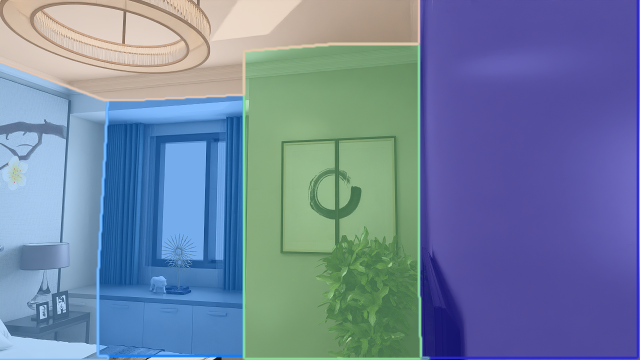} &
		\includegraphics[width=0.15\linewidth]{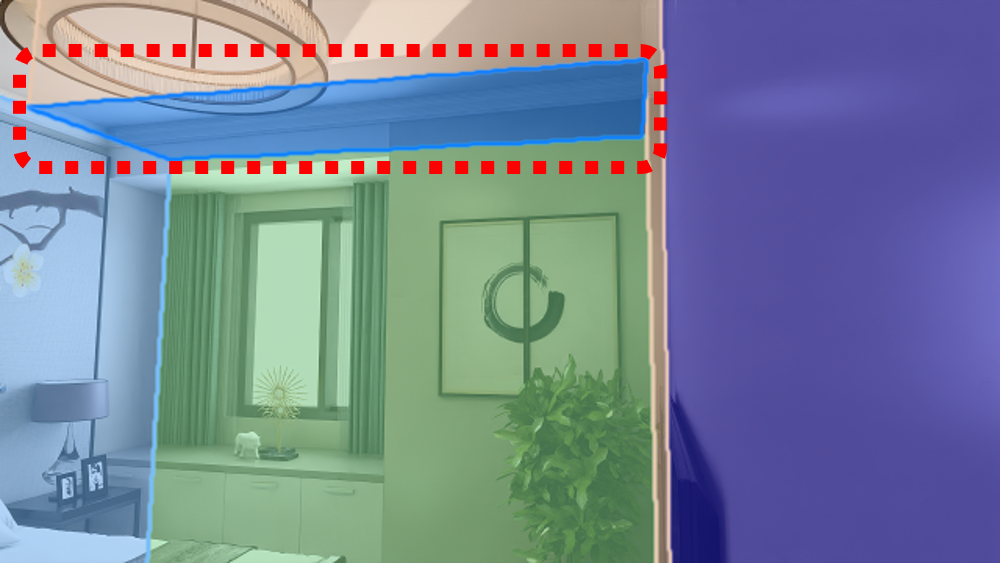} &
		\includegraphics[width=0.15\linewidth]{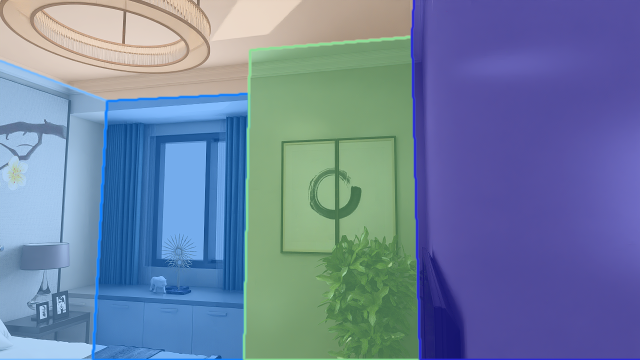} &
		\includegraphics[width=0.15\linewidth]{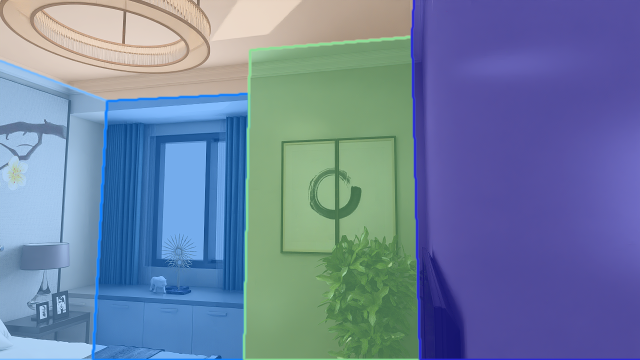} &
		\includegraphics[width=0.15\linewidth]{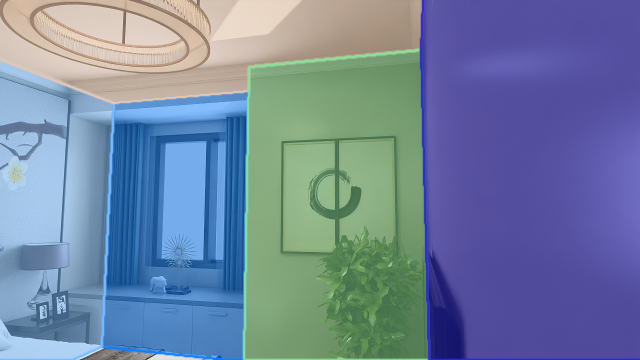}
		\tabularnewline
		\includegraphics[width=0.15\linewidth]{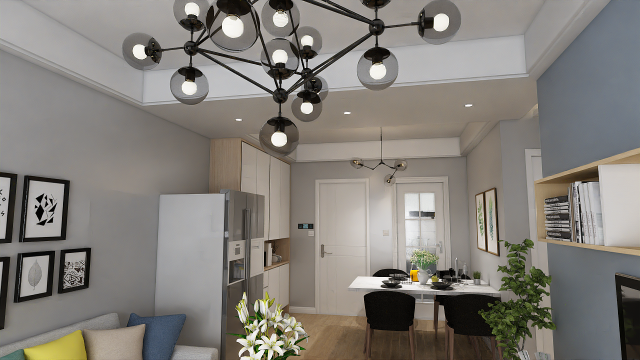} &
		\includegraphics[width=0.15\linewidth]{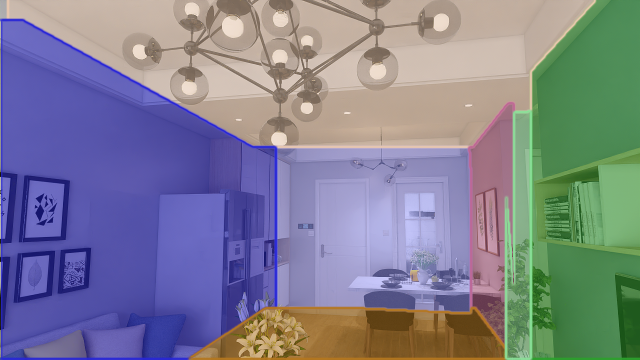} &
		\includegraphics[width=0.15\linewidth]{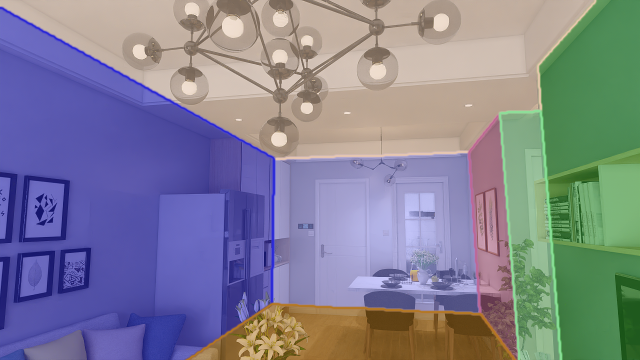} &
		\includegraphics[width=0.15\linewidth]{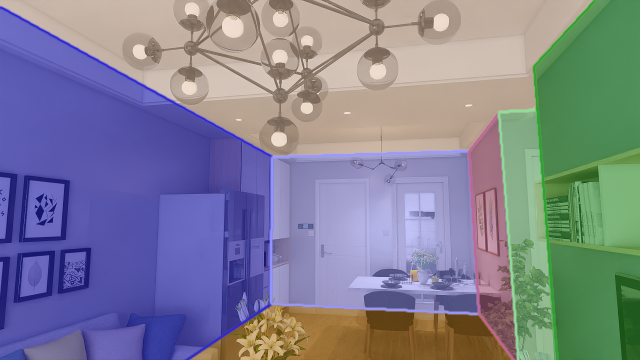} &
		\includegraphics[width=0.15\linewidth]{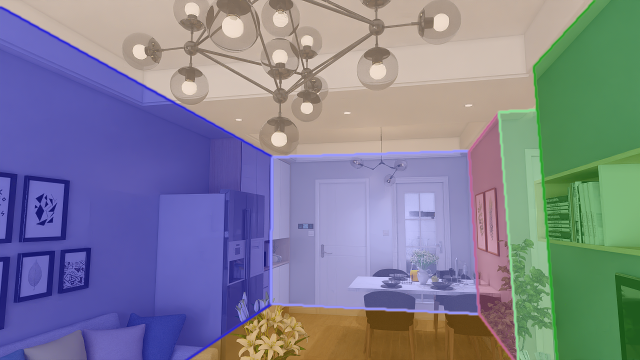} &
		\includegraphics[width=0.15\linewidth]{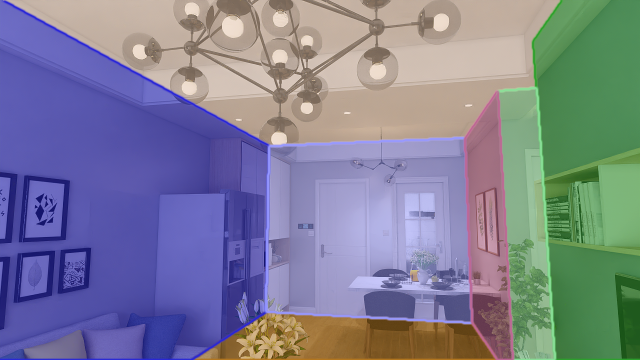}
		\tabularnewline
		Input image & Planar R-CNN & RaC & {\footnotesize Ours (w/o optimization)} & Ours & Ground truth
		\tabularnewline
	\end{tabular}
	\caption{Qualitative results on Structured3D dataset~\cite{ZhengZLTGZ20}. The correspondences are marked in a common color. We highlight the major differences in the dashed red bounding boxes.}
	\label{fig:s3d:2d}
\end{figure*}

\PAR{Evaluation metric.} Following RaC~\cite{StekovicFL20}, we adopt four standard evaluation metrics: (i) IoU: intersection over the union between the predicted plane layout and the ground truth, (ii) Pixel Error (PE): pixel-wise error between predicted 2D plane segmentation and the ground truth, (iii) Edge Error (EE): the symmetrical Chamfer distance between predicted layout boundary and the ground truth, (iv) Root Mean Square Error (RMSE) between the predicted layout depth map and the ground truth. For the 2D metrics (\ie, IoU and PE), we match the predicted plane segmentation to the ground truths. Starting from the largest ground-truth segmentation, we iteratively find the predicted segmentation with the highest IoU score. Each ground truth is only allowed to be matched at most once.

\PAR{Quantitative evaluation.} \tabref{tab:s3d} shows the quantitative results and runtime of all methods. All times are measured on the same computing platform with an Intel Xeon Gold 6128 @ 3.4GHz (24 cores) and a single NVIDIA TITAN Xp GPU. As one can see, our proposed method achieves state-of-the-art performance without optimization. Compared with RaC~\cite{StekovicFL20}, our method reasons the connectivity relations between planes and handles with occlusions through introducing room layout assumption and vertical line detection, and avoids complex optimization as well as decreasing the running time from \SI{5.35}{\second} to \SI{0.07}{\second}. Furthermore, when using the proposed layout optimization algorithm to reconstruct a geometrically consistent room layout between planes and lines, our method provides a good trade-off between accuracy and speed. This clearly demonstrates the effectiveness of the proposed method.

\PAR{Qualitative evaluation.} \figref{fig:s3d:3d} shows our 3D room layout reconstruction results for a variety of scenes. The qualitative comparisons against existing non-cuboid room layout estimation methods show in \figref{fig:s3d:2d}. We make the following observations: (i) All methods perform well in simple scenarios (\eg, the first row). (ii) Since we explicitly use detected lines as geometric cues to optimize the room layout, our results can preserve the boundary of two adjacent walls well (\eg, the second and third row). (iii) Planar R-CNN does not reason how the adjacent walls connect and bounds each plane segmentation by their bounding boxes. As a result, two adjacent walls may inter-penetrate each other (\eg, the third row) or do not touch each other (\eg, the fourth row). (iv) The carefully designed heuristics of RaC are not robust in every configuration (\eg, the fifth row). Furthermore, when the number of plane candidates is large, RaC is very computationally expensive (\eg, about \num{8} minutes for the sixth row).

\begin{figure}[t]
	\centering
	\setlength{\tabcolsep}{1pt}
	\begin{tabular}{ccc}
		\includegraphics[width=0.3\linewidth]{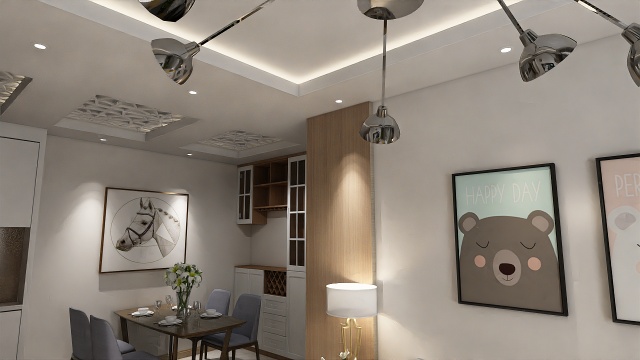} &
		\includegraphics[width=0.3\linewidth]{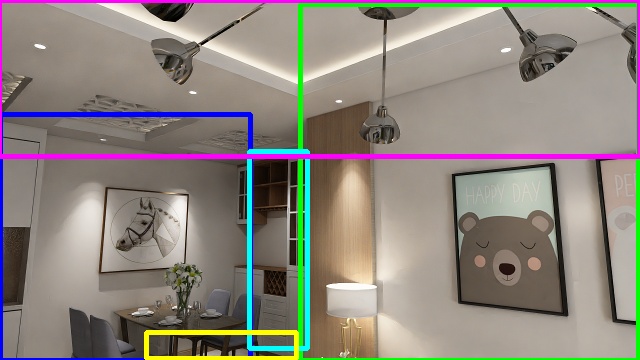} &
		\includegraphics[width=0.3\linewidth]{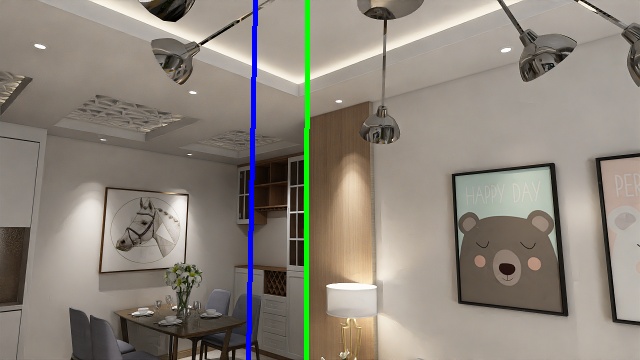}
		\tabularnewline
		\includegraphics[width=0.3\linewidth]{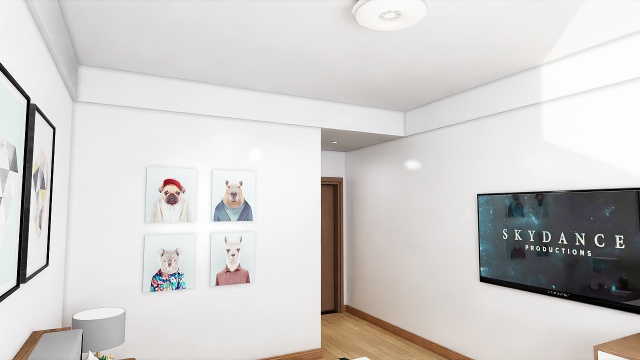} &
		\includegraphics[width=0.3\linewidth]{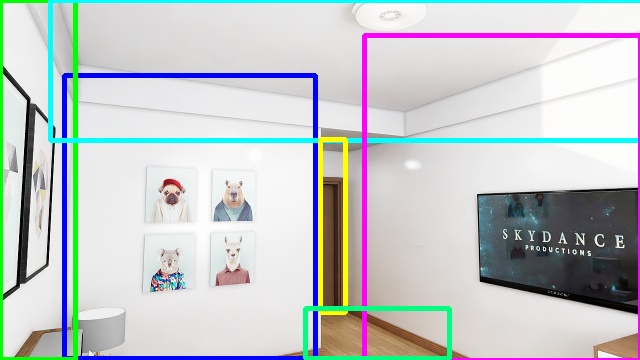} &
		\includegraphics[width=0.3\linewidth]{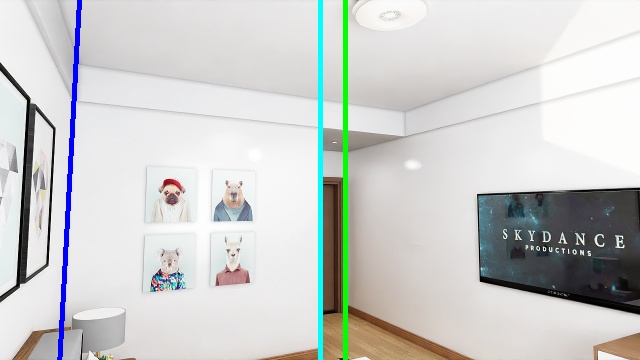}
		\tabularnewline
		\includegraphics[width=0.3\linewidth]{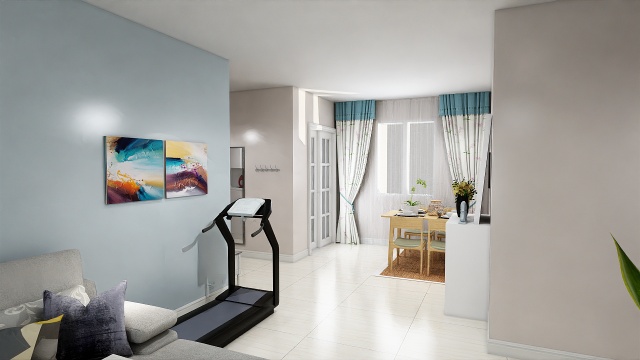} &
		\includegraphics[width=0.3\linewidth]{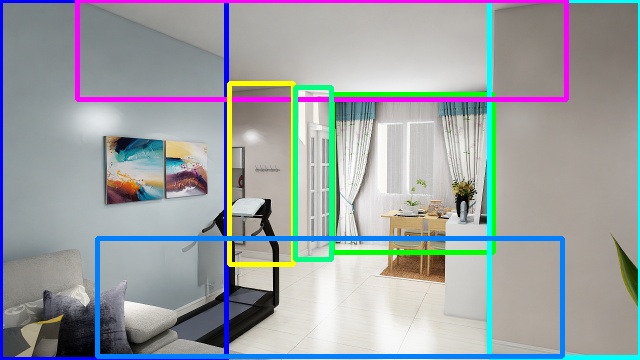} &
		\includegraphics[width=0.3\linewidth]{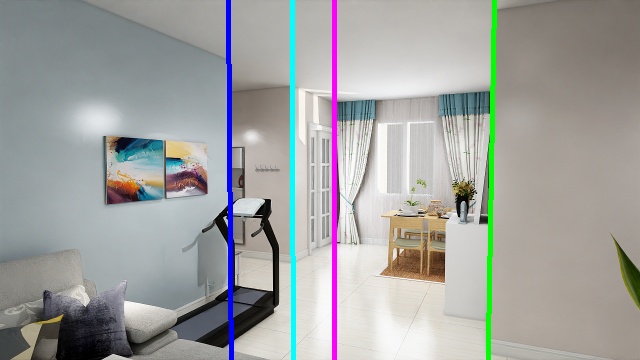}
		\tabularnewline
		\includegraphics[width=0.3\linewidth]{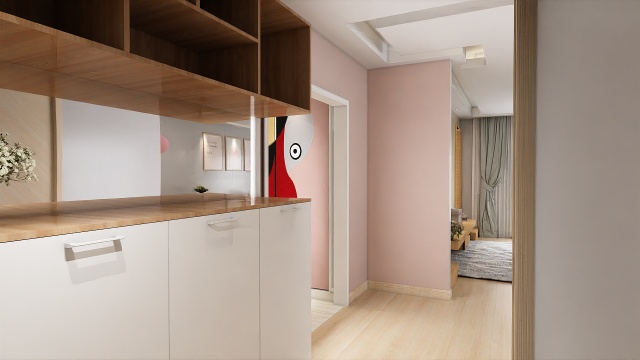} &
		\includegraphics[width=0.3\linewidth]{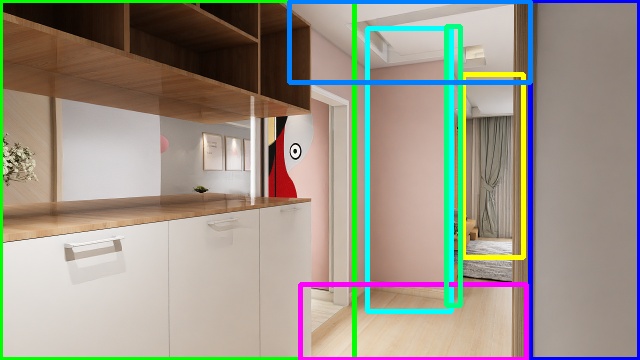} &
		\includegraphics[width=0.3\linewidth]{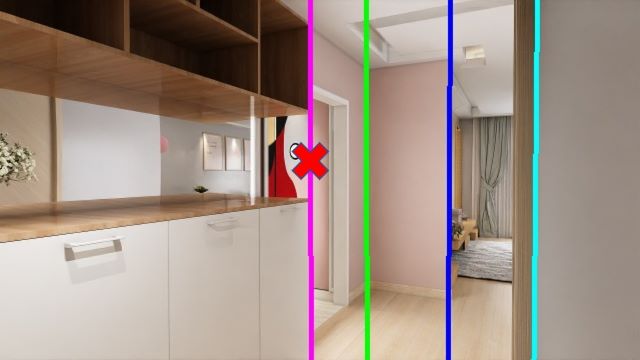}
		\tabularnewline
		\includegraphics[width=0.3\linewidth]{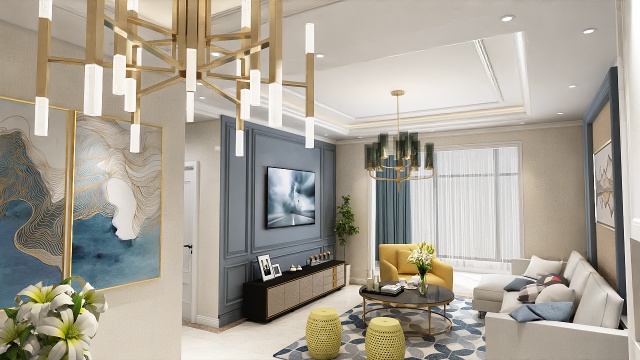} &
		\includegraphics[width=0.3\linewidth]{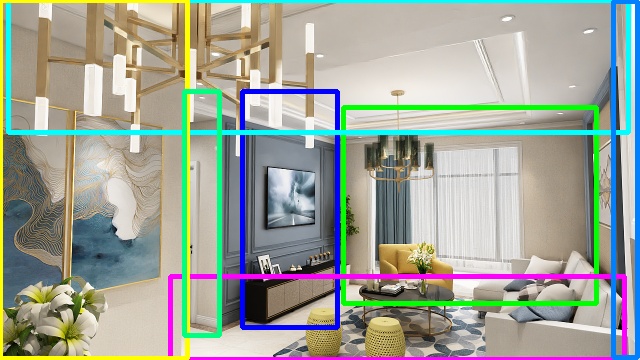} &
		\includegraphics[width=0.3\linewidth]{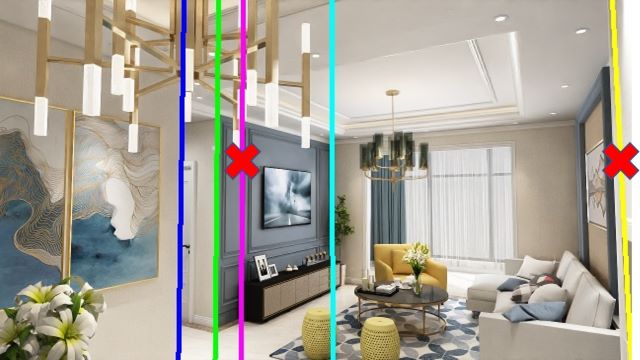}
		\tabularnewline
		Input image & Plane detections & Line detections
		\tabularnewline
	\end{tabular}
	\caption{Plane detections and vertical lines detections between adjacent walls on Structured3D dataset~\cite{ZhengZLTGZ20}. The lines with red cross are filtered by the potential intersection line region.}
	\label{fig:s3d:detections}
\end{figure}

\figref{fig:s3d:detections} shows our detection results of planes and vertical lines between adjacent walls. Thanks to the advent of powerful detection technology, we can accurately detect planes and lines. Meanwhile, we only consider the lines that locate in the potential intersection line region, which reduces the false positives, such as texture lines.

\subsection{Results on NYUv2 303 Dataset}

We further evaluate the performance of our approach on a real but much smaller dataset, \ie, NYUv2 303 dataset~\cite{ZhangKSU13}, which contains \num{303} images from NYUv2 dataset~\cite{SilbermanHKF12}. The resolution of input image is $640 \times 480$. Note that the dataset only provides cuboid-based layout annotation (hence evaluation on this dataset may favor cuboid-based methods).

\begin{figure}[t]
	\centering
	\setlength{\tabcolsep}{1pt}
	\begin{tabular}{ccc}
		\includegraphics[width=0.3\linewidth]{figures/nyu/img/22.png} &
		\includegraphics[width=0.3\linewidth]{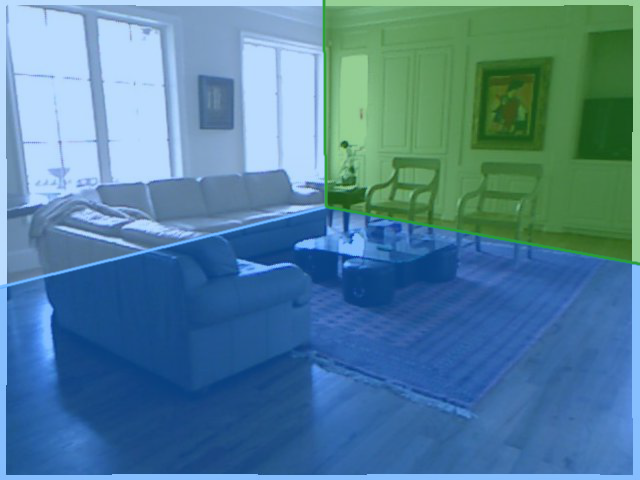} &
		\includegraphics[width=0.3\linewidth]{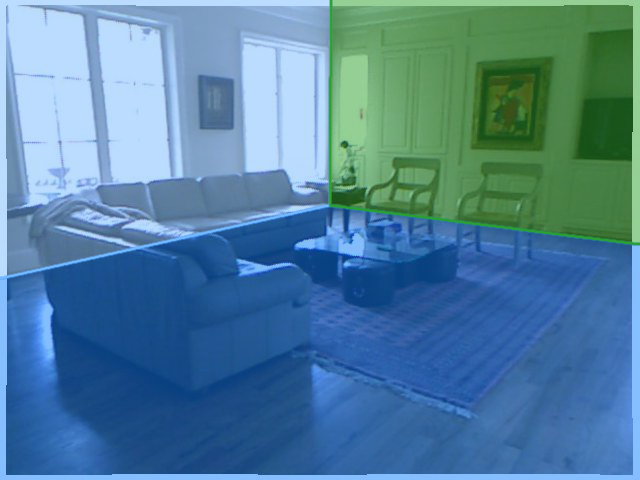}
		\tabularnewline
		\includegraphics[width=0.3\linewidth]{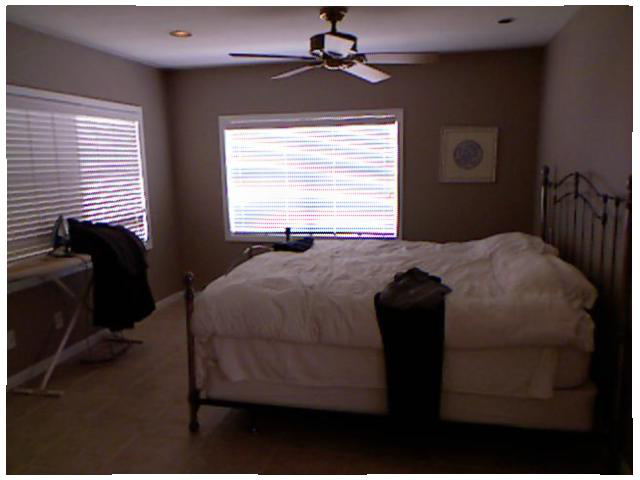} &
		\includegraphics[width=0.3\linewidth]{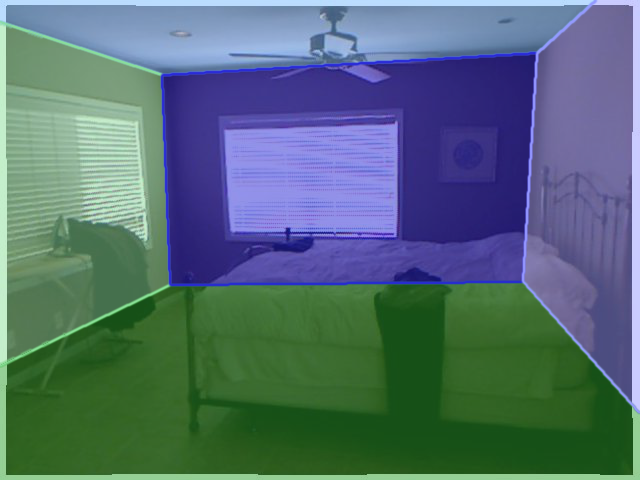} &
		\includegraphics[width=0.3\linewidth]{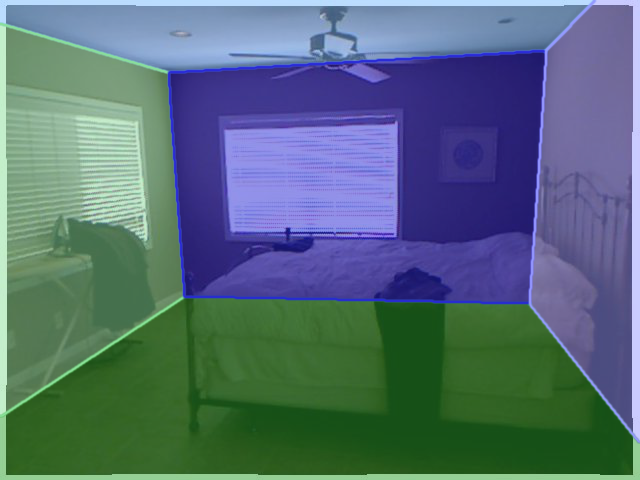}
		\tabularnewline
		\includegraphics[width=0.3\linewidth]{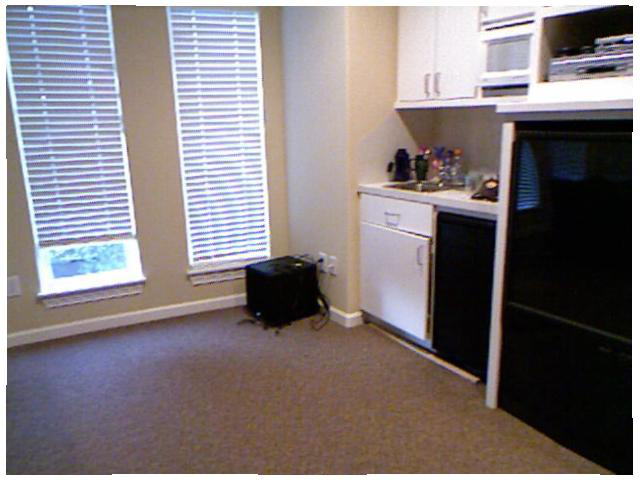} &
		\includegraphics[width=0.3\linewidth]{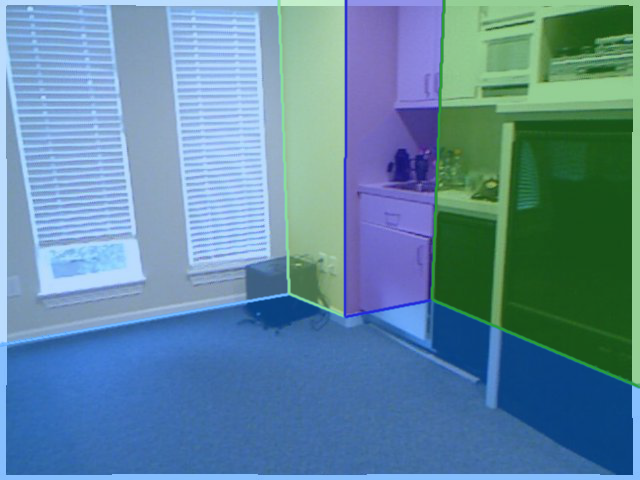} &
		\includegraphics[width=0.3\linewidth]{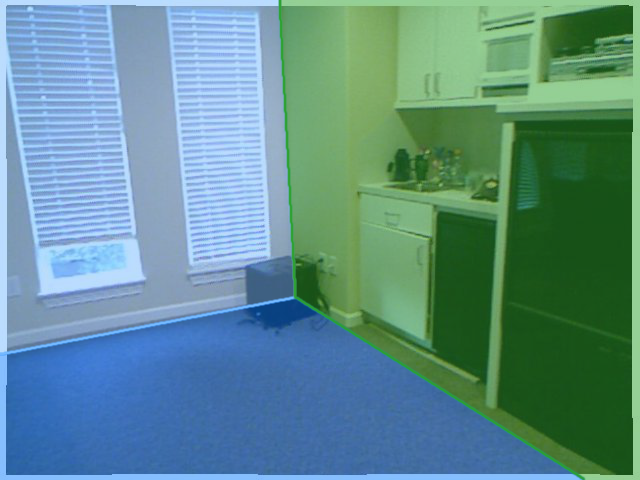}
		\tabularnewline
		\includegraphics[width=0.3\linewidth]{figures/nyu/img/46.png} &
		\includegraphics[width=0.3\linewidth]{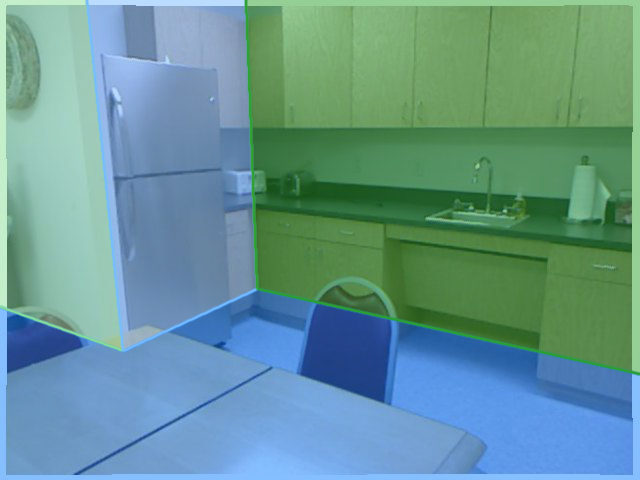} &
		\includegraphics[width=0.3\linewidth]{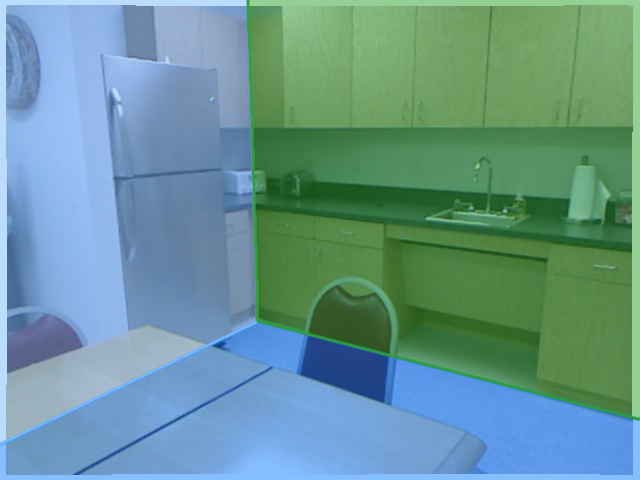}
		\tabularnewline
		Input image & Ours & Ground truth
		\tabularnewline
	\end{tabular}
	\caption{Qualitative results on NYU 303 dataset~\cite{ZhangKSU13}. The correspondences are marked in a common color.}
	\label{fig:nyu}
\end{figure}

\begin{table}[t]
	\centering
	\sisetup{
		table-number-alignment 	= center,
		table-figures-integer 	= 2,
		table-figures-decimal 	= 2,
		table-auto-round 		= true,
		detect-weight			= true,
	}
	\begin{tabular}{lS}
		\toprule
		{Method} & {PE (\%) $\downarrow$} \tabularnewline
		\midrule
		\citet{SchwingHPU12} & 13.66 \tabularnewline
		\citet{ZhangKSU13} & 13.94 \tabularnewline
		RoomNet~\cite{LeeBMR17}$^\dagger$ & 12.31 \tabularnewline
		PlaneNet~\cite{LiuYCYF18} & 12.64 \tabularnewline
		\citet{HirzerRL20} & \bfseries 8.49 \tabularnewline
		\midrule
		Planar R-CNN~\cite{HowardJLP18} & 12.19 \tabularnewline
		RaC~\cite{StekovicFL20} & 13.0 \tabularnewline
		Ours (w/o optimization) & 11.245394 \tabularnewline
		Ours & \bfseries 10.606974 \tabularnewline
		\bottomrule
	\end{tabular}
	\caption{Quantitative results on NYUv2 303 dataset~\cite{ZhangKSU13}. $\dagger$: The results are reported by \cite{HirzerRL20}.}
	\label{tab:nyu}
\end{table}

\PAR{Quantitative evaluation.} \tabref{tab:nyu} shows the quantitative comparisons against both cuboid-based and non-cuboid methods on the NYUv2 303 dataset. To validate our approach, we follow two recent non-cuboid methods~\cite{HowardJLP18, StekovicFL20} and use the Hungarian algorithm to match detected planes to the ground truths. As one can see, our methods outperform  all non-cuboid layout methods and almost all cuboid-based methods (with the exception of \citet{HirzerRL20}). Nevertheless, such cuboid-based methods cannot be applied to the more extensive and flexible Structured3D dataset.

\PAR{Qualitative evaluation.} \figref{fig:nyu} shows room layout estimation results on the NYU 303 dataset. The first two examples show that our approach can predict cuboid-shape layout even without such an assumption. The last two examples demonstrate that our method can predict the more refined correct layouts than the ground truth (cuboids).

\subsection{Failure Cases}

\begin{figure}[t]
	\centering
	\setlength{\tabcolsep}{1pt}
	\begin{tabular}{ccc}
		\includegraphics[width=0.3\linewidth]{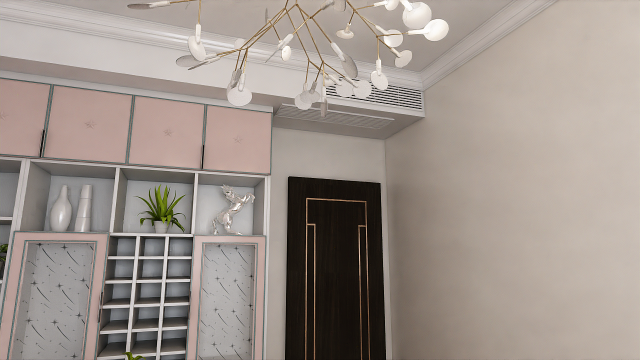} &
		\includegraphics[width=0.3\linewidth]{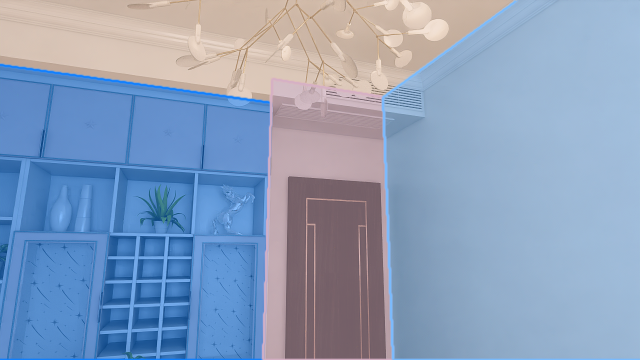} &
		\includegraphics[width=0.3\linewidth]{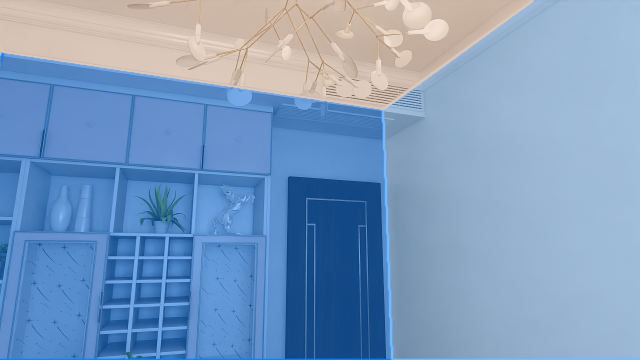}
		\tabularnewline
		\includegraphics[width=0.3\linewidth]{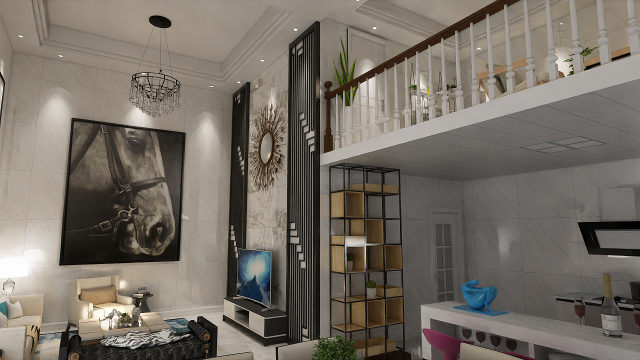} &
		\includegraphics[width=0.3\linewidth]{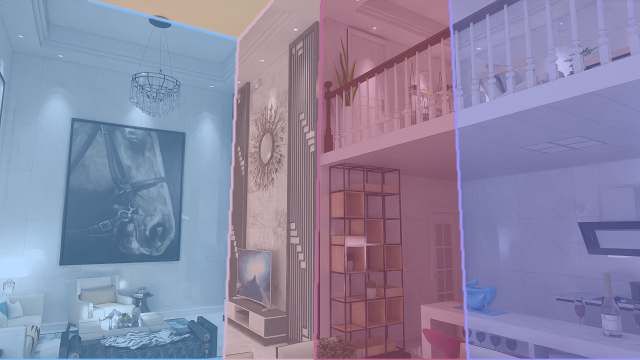} &
		\includegraphics[width=0.3\linewidth]{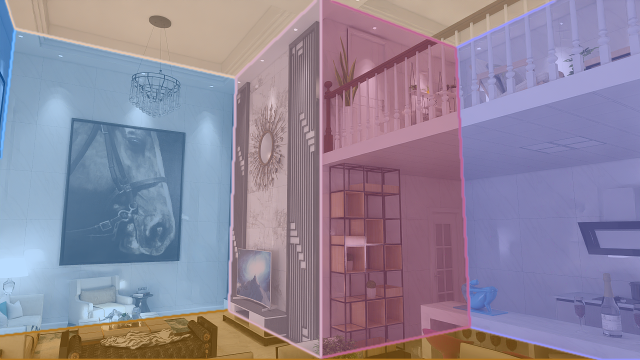}
		\tabularnewline
		\includegraphics[width=0.3\linewidth]{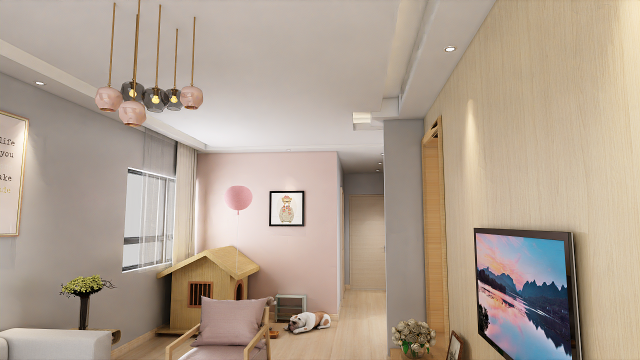} &
		\includegraphics[width=0.3\linewidth]{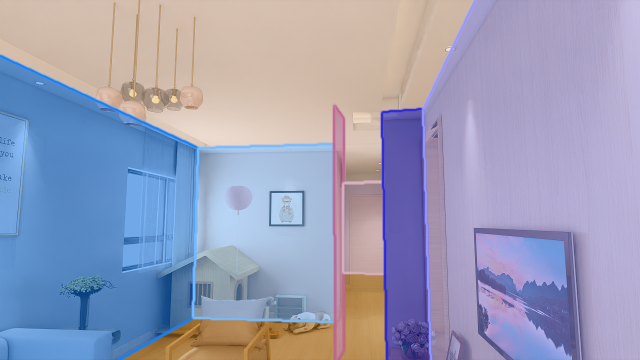} &
		\includegraphics[width=0.3\linewidth]{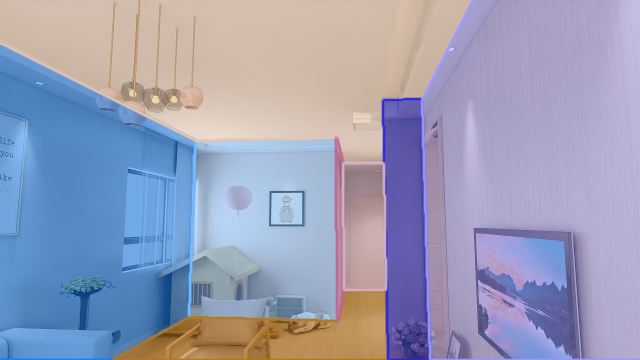}
		\tabularnewline
		Input image & Ours & Ground truth \tabularnewline
	\end{tabular}
	\caption{Failure Cases. The correspondences are marked in a common color.}
	\label{fig:failure}
\end{figure}

We show some failure cases of our method in \figref{fig:failure}. In the first example, our method mistakenly detects the foreground furniture as the wall. A possible reason is that most of the wall is occluded by the foreground furniture. In the second example, our approach fails to detect the floor, which leads to an inaccurate wall-floor boundary. In the third example, the ceiling-wall and floor-wall boundaries of the light pink plane are not precisely localized due to the inaccurate 3D parameter estimations of the small plane.

\section{Conclusion}

This paper proposes a simple yet effective 3D room layout reconstruction approach assuming that the room layout consists of a single ceiling, a single floor, and several vertical walls. Specifically, we first employ CNNs to detect 3D planes and vertical lines. Then, we adopt a geometric reasoning method to achieve the room layout reconstruction. Finally, to reconstruct a geometrically consistent layout, we optimize the 3D parameters of the planes to align detected lines. The proposed method achieves state-of-the-art performance on the largest non-cuboid room layout benchmark. In the future, we will consider introducing more geometric cues, such as wall-ceiling/wall-floor intersection lines, and relax the assumption to handle multi-tiered ceilings/floors.

\PAR{Acknowledgements.} We would like to thank Zihan Zhou and Weixin Luo for valuable comments on this paper.

{
\small
\bibliographystyle{ieee_fullname}
\bibliography{references}
}

\end{document}